\documentclass[graybox]{svmult}

\usepackage{type1cm}        
\usepackage{makeidx}         
\usepackage{graphicx}        
\usepackage{multicol}        
\usepackage[bottom]{footmisc}

\usepackage{cite}
\usepackage{algorithmic}
\usepackage{algorithm}
\usepackage{graphicx}
\usepackage{xcolor}
\usepackage{mathtools}
\usepackage{tabularx}
\usepackage{makecell}
\usepackage{dsfont}
\usepackage[absolute,overlay]{textpos}

\usepackage{todonotes} 

\usepackage{newtxtext}       %
\usepackage[varvw]{newtxmath}       

\makeindex             

\def\etal{\emph{~et~al. }}
\DeclarePairedDelimiter\abs{\lvert}{\rvert}

\begin{document}

\title{Learning-based social coordination to improve safety and robustness of cooperative autonomous vehicles in mixed traffic}

\author{Rodolfo Valiente, Behrad Toghi, Mahdi Razzaghpour, Ramtin Pedarsani, Yaser P. Fallah}

\maketitle

\abstract{It is expected that autonomous vehicles(AVs) and heterogeneous human-driven vehicles(HVs) will coexist on the same road. The safety and reliability of AVs will depend on their social awareness and their ability to engage in complex social interactions in a socially accepted manner. However, AVs are still inefficient in terms of cooperating with HVs and struggle to understand and adapt to human behavior, which is particularly challenging in mixed autonomy. 
In a road shared by AVs and HVs, the social preferences or individual traits of HVs are unknown to the AVs and different from AVs, which are expected to follow a policy, HVs are particularly difficult to forecast since they do not necessarily follow a stationary policy. To address these challenges, we frame the mixed-autonomy problem as a multi-agent reinforcement learning (MARL) problem and propose an approach that allows AVs to learn the decision-making of HVs implicitly from experience, account for all vehicles' interests, and safely adapt to other traffic situations. In contrast with existing works, we quantify AVs' social preferences and propose a distributed reward structure that introduces altruism into their decision-making process, allowing the altruistic AVs to learn to establish coalitions and influence the behavior of HVs. 
}

\section{Introduction}
\label{sec:1}
Autonomous vehicles (AVs) have been an attractive research area for decades, as it offers the potential to generate more efficient and safer road networks~\cite{cosgun2017towards}. The adoption of AVs will not become a reality until they can co-exist with humans, as part of a complex social system. In order to maximize the potential of AVs and optimize for safety and traffic efficiency of all the vehicles on the road, AVs have to coordinate and influence the other agents~\cite{cosgun2017towards,schwarting2019social,sagberg2015review}. 

We recognize the importance of social interaction and behavior in safety and reliability and identify two important research directions. First, AVs must be social actors and behave predictably and safely. 
Driver behavior is shaped by habits and expectations in the traffic environment. The vehicle's interaction will be influenced by the way AV decisions are perceived. Therefore, the ability of AVs to drive in a socially obedient manner is critical for the safety of passengers and other vehicles because predictable behavior allows humans to comprehend and respond appropriately to the AV's actions. Second, AVs must be social-aware and learn to identify social cues of egoism or altruism, understand the behavior of human drivers and learn how to interact and coordinate with all agents in a mixed traffic environment, adapting and influencing the HVs behaviors to optimize for a social utility that improves traffic flow and safety.

\begin{figure*}[t!]
\centering
\includegraphics[width=.99\textwidth]{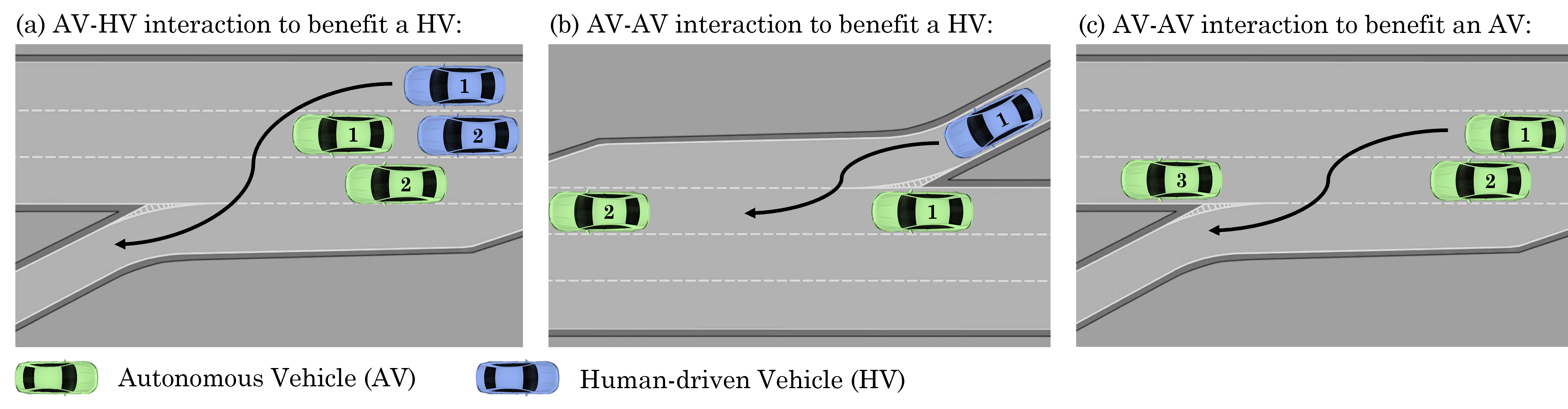}
\caption{\small{
\textbf{(a) Interaction of AV-HV to benefit a HV:} Altruistic agents create alliances and direct the behavior of HVs to improve traffic flow and prevent dangerous circumstances.  AV1 and AV2 can create a formation to guide HV2 and provide a route for HV1, allowing the HV to change lanes and navigate to the exit ramp.
\textbf{(b) Interaction of AV-AV to benefit a HV:} The goal of HV1 is to integrate onto the highway. Egoistic AVs disregard the merging vehicle and do not make room for it, possibly resulting in dangerous situations, however, if they exhibit sympathy for the merging HV, they can compromise on their own interest to create a safe path for HV1 to merge into the highway.
\textbf{(c) Interaction of AV-AV to benefit another AV:}  The goal of AV1 is to exit the highway. If AV2 acts selfishly, AV1 may miss the exit and be unable to complete its task. However, if AV2 and AV3 consider AV1's mission and act altruistically, they can free up space in the platoon by AV2 decelerating and AV3 accelerating, allowing AV1 to safely take the exit.
}}
\label{fig:generalfigure}
\end{figure*}

In this chapter, we focus on social awareness challenges and seek a solution that can ensure the safety and robustness of AVs in the presence of human drivers with heterogeneous behavioral traits. Vehicle-to-vehicle (V2V) communication allows connected and autonomous vehicles (CAVs) to interact directly with their neighbors~\cite{toghi2019spatio, shah2019real}. By using V2V communication CAVs can create an extended perception that facilitates explicit cooperation among vehicles to overcome the limits of a non-cooperative agent~\cite{valiente2019controlling, razzaghpour2021impact, valiente2020dynamic}. While planning in a fully AV scenario is relatively easy to achieve, coordination in the presence of HVs is a significantly more challenging task, as the AVs not only need to react to road objects but also need to consider the behaviors of HVs~\cite{aoki2020cooperative, toghi2021cooperative, Jami2022AugmentedDB}. 

In contrast to the individual non-cooperative approaches, we investigate the mixed-autonomy decision-making challenge from a multi-agent and social perspective. Rewarding AVs for adopting an altruistic behavior and taking into consideration the interests of other vehicles allows them to see the broad picture and find solutions that maximize the utility of the group. In addition to the potential benefits of altruistic decision-making in terms of safety and efficiency, altruism results in more societally advantageous outcomes~\cite{schwarting2019social}. Figure~\ref{fig:generalfigure}(a) demonstrates how a group of AVs can guide HV to increase safety and efficiency, while Figure~\ref{fig:generalfigure}(b) and~\ref{fig:generalfigure}(c) show how AVs can collaborate to accomplish a social objective that benefits another HV or AV.

Currently, AVs lack an understanding of human behavior and frequently act extremely cautiously to avoid collisions. This conservative behavior not only leaves AVs unprotected from aggressive HVs, but also results in unexpected reactions that confuse HVs, creating bottlenecks in traffic flow and causing accidents. 
It's critical to distinguish between a human driver's individual traits, such as aggressiveness, conservativeness, and risk tolerance, and their social preferences, such as egoism and altruism. Even though the two categories are related, they have distinctive natures and so behave differently in mixed traffic. An aggressive driver, for example, is not inherently egoistic or selfish, but their aggression may hinder their ability to collaborate with other drivers and participate in a socially desirable coexistence with AVs~\cite{sagberg2015review,harris2014prosocial,vallieres2014intentionality}. In the field of behavior planning for AVs in mixed-autonomy traffic, we identify two fundamental problems. First, human drivers differ considerably in their individual traits and social preferences, making AV behavior planning exceedingly difficult because it is difficult for the AV to foresee the type of behavior it would encounter when dealing with a human driver. Furthermore, relying on a real-time inference of HV behaviors is not always feasible because vehicle interactions can be brief, such as when two vehicles meet at an intersection. Second, driving requires complex interactions of agents in a partially observable and non-stationary environment, as HVs do not follow a fixed policy and modify their policies in real-time in response to the actions of other vehicles.

\begin{figure}[b]
  \centering
  \includegraphics[width=.7\textwidth]{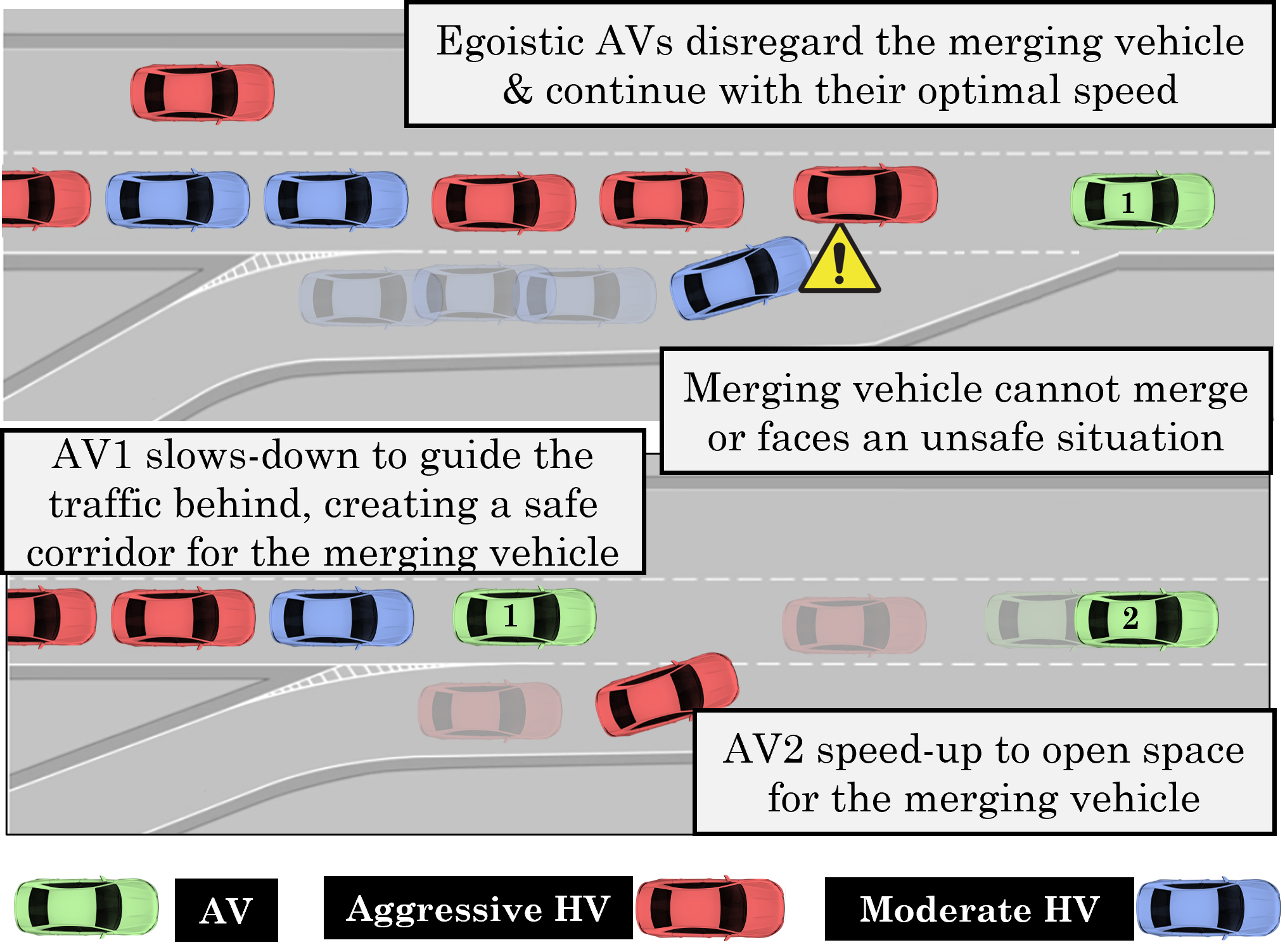}
  \caption{\small{For a seamless and safe highway merging, all AVs must coordinate and account for the utility of HVs. \emph{(top)} Egoistic AVs optimize only for their own utility, \emph{(bottom)} Altruistic AVs consider also the HV's utility.}}
\label{fig:mainfigure}
\end{figure}

\begin{figure}[b]
  \centering
  \includegraphics[width=.7\textwidth]{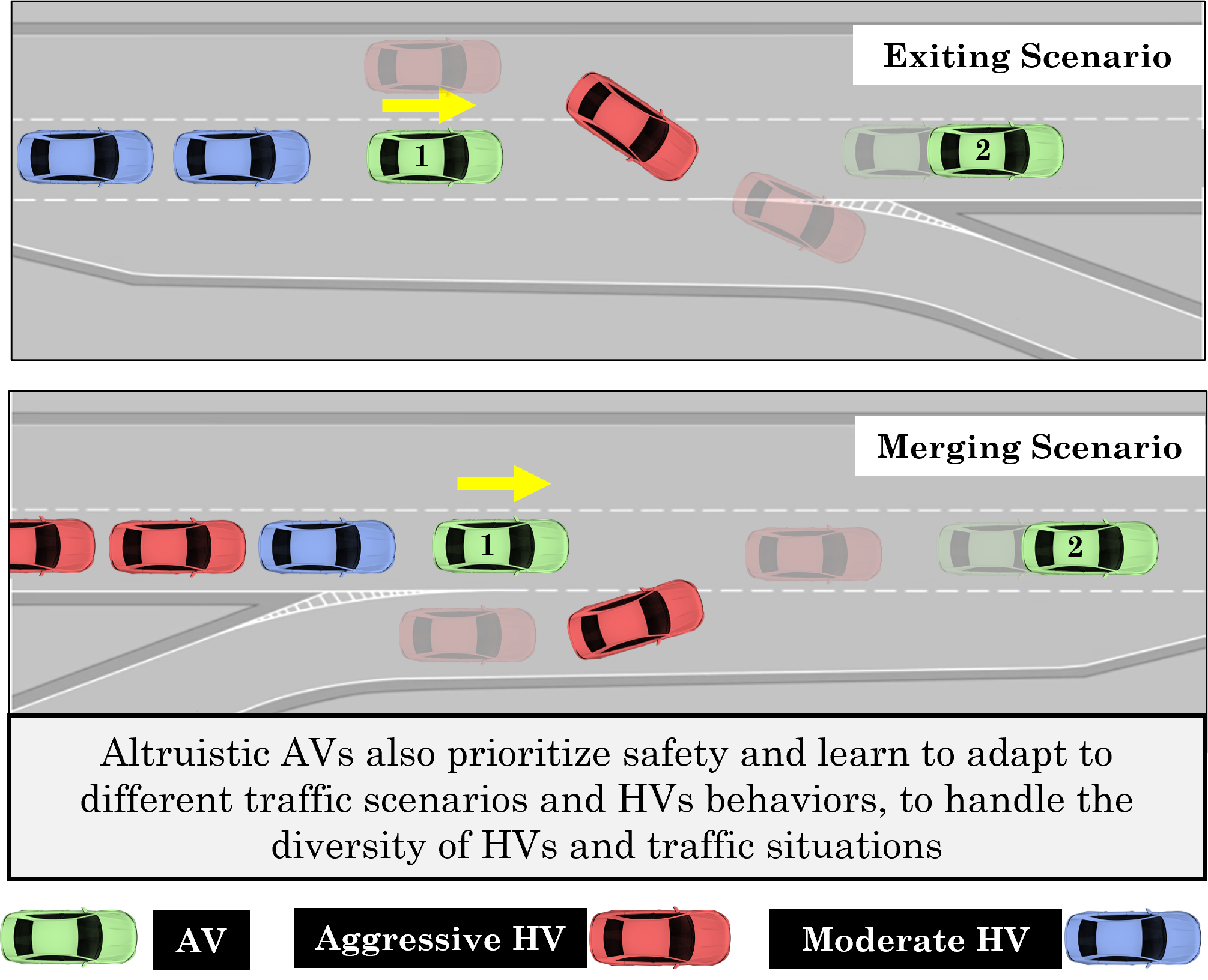}
  \caption{\small{Highway exiting and merging scenarios with AVs (green) and aggressive HVs (red) sharing the road. Altruistic AVs must learn to cooperate to exit/merge successfully and safely while being adaptable to a variety of scenarios and HV behaviors.}}
\label{fig:mainfigure2}
\end{figure}

The integration of AVs into the real world requires them to address those challenges. Due to the differences in maneuverability and reaction time between AVs and HVs, a road shared by both becomes a competitive situation. In contrast with the full-autonomy case, here the coordination between HVs and AVs is not as straightforward since AVs do not have an explicit means of harmonizing with humans and are therefore required to locally account for the other HVs and AVs in their proximity.
This dilemma intensifies if AVs act egoistically and optimize solely for their local utility. As an illustration, Figure~\ref{fig:mainfigure} and Figure~\ref{fig:mainfigure2} demonstrate a highway exiting and merging scenario in mixed traffic. We consider a general setup where AVs and HVs with different behaviors coexist. Vehicles need to efficiently merge onto the lane or exit the highway without collisions. In an ideal cooperative environment, AVs should proactively decelerate or accelerate to provide sufficient room for vehicles to safely exit/merge and prevent hazardous situations, while also being resilient to various situations and behaviors and assuring safety in decision-making~\cite{bouton2019cooperation}. For instance, in Figure~\ref{fig:mainfigure} (merging scenario) if AVs act egotistically, the merging vehicle must rely on the HV to slow down to allow it to merge. However, due to the unpredictability of HVs, relying solely on HVs might result in suboptimal or even dangerous circumstances. Therefore, if all AVs act egotistically, the merging vehicle would either be unable to join the highway or it will wait for an HV and risk cutting into the highway without knowing whether the HV will slow down. Nevertheless, if AVs act altruistically, they can coordinate to guide the traffic on the highway to allow for a seamless and safe merging. In particularly challenging driving scenarios, such altruistic AVs can achieve societally beneficial results without relying on or making assumptions about HVs behaviors.

To address these challenges, existing literature either depend on models of human behavior generated from pre-recorded driving datasets~\cite{sadigh2016planning, wu2017emergent} or define social utilities that can impose cooperative behavior among AVs and HVs~\cite{toghi2021social}. Other works focus on rule-based methods that use heuristics and hand-coded rules to guide the AVs~\cite{rios2016survey} or probabilistic driver modeling~\cite{mahjoub2019representing} learned from human driving data. While this is feasible for simple situations, these methods become impractical in complex scenarios. Additionally, the human driver models learned in the absence of AVs, are not necessarily valid when humans confront AVs. This limits the application of the generated solutions, as they are frequently limited to the human behaviors with which AVs interacted during training. To account for this, several works in the literature adopt an excessively cautious approach when interacting with humans~\cite{li2018safe}. This strategy not only leaves the AVs vulnerable to other aggressive drivers, particularly in competitive situations, but it also causes traffic congestion and significant safety risks~\cite{cosgun2017towards, schwarting2019social}.

On the other hand, data-driven methods such as reinforcement learning (RL) have received increased attention~\cite{lin2020anti} as RL-based methods can learn decision-making and driving behaviors that are hard for traditional rule-based designs. However, the majority of the RL approaches are designed for a single AV, or try to handle the interaction between AVs and HVs either by predicting human behavior or by relying on the fact that humans are willing to collaborate or can be influenced to do so~\cite{sadigh2016planning,sadigh2018planning}, which could compromise safety or lead to sub-optimal performance. Recent works consider social interactions of AVs and train altruistic AVs that learn from experience and influence HVs to optimize a social utility function that benefits all vehicles on the road~\cite{toghi2021altruistic,toghi2021cooperative}.

In contrast, we consider a data-driven multi-agent reinforcement learning (MARL) approach and let the autonomous agents implicitly learn the decision-making process of human drivers only from experience, while optimizing for a social utility. By incorporating a cooperative reward structure into our MARL framework, we can train AVs that coordinate with each other, sympathize with HVs, and, as a result, demonstrate enhanced performance in competitive driving scenarios, such as highway exiting and merging. Despite not having access to an explicit model of the human drivers, the trained autonomous agents learn to implicitly model the environment dynamics, including the behavior of human drivers, which enables them to interact with HVs and guide their behavior.

\begin{figure*}[t]
  \centering
  \includegraphics[width=.99\textwidth]{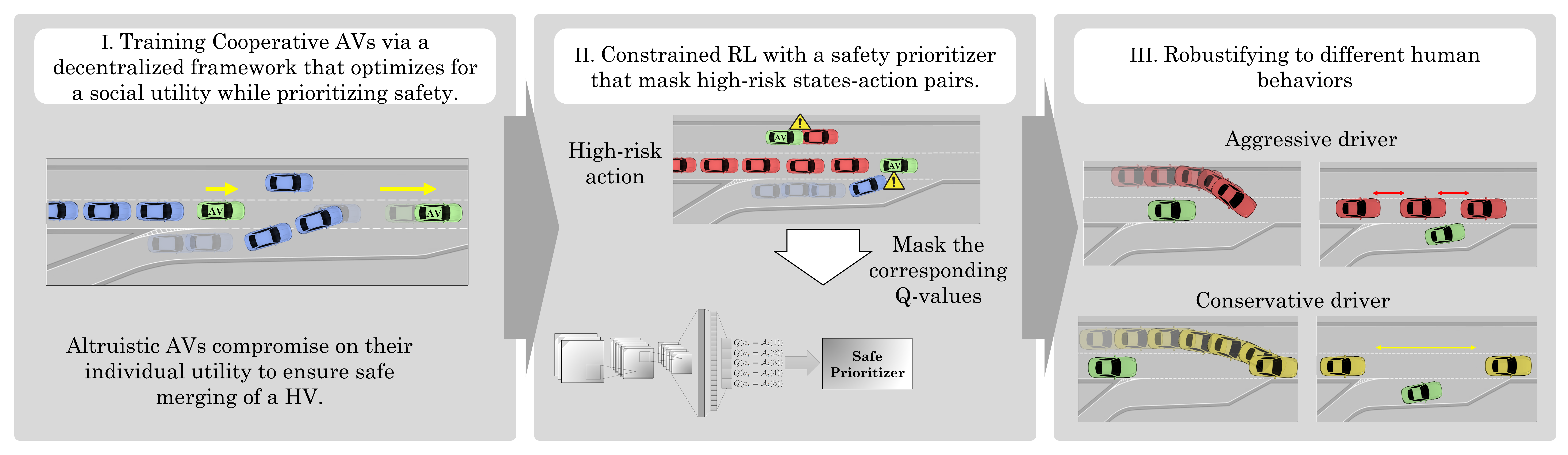}
  \caption{\small{An overview of our approach to leverage social awareness and coordination to improve the safety and reliability of CAVs. Our social-aware AVs learn from scratch not only to drive but also to understand the behavior of HVs and coordinate with them, they learn to adapt and influence HVs in a robust and safe manner.}}
  \label{fig:cover}
\end{figure*}

This research aims to create a safe and robust training regimen that allows AVs to collaborate and influence the behavior of human drivers to achieve socially desirable outcomes, regardless of HV individual traits and social preferences. 
We based our work on the following insights.
First, we rely on a decentralized reinforcement learning architecture that optimizes for a social utility that learns from experience and exposes the learning agents to a wide range of driving. As a result, the agents become more resistant to human driver behavior and can handle cooperative-competitive behaviors regardless of HV's hostility or social preference. Second, a safety prioritizer is presented to minimize high-risk actions that could jeopardize driving safety. The safety prioritizer constrains the policy of cooperative AVs to ensure the safety of their behavior via masking the Q-states that lead to high-risk outcomes. Figure~\ref{fig:cover} shows an overview of our process.

Our main contributions are summarized as follows:
\begin{itemize}
    \item We formulate the mixed-autonomy problem as a decentralized MARL problem and present an approach to training altruistic agents which utilizes a decentralized reward mechanism for achieving socially advantageous behaviors and takes advantage of a 3D convolutional deep reinforcement learning architecture to capture the temporal information in driving data.
  \item A training algorithm is proposed to make AVs robust to different drivers' behavior and situations while producing socially desirable outcomes. We investigate the effect of HVs behaviors on our altruistic AVs agents and especially conclude that the higher the traffic aggressiveness, the higher the importance of social coordination.
\item We investigate the scenarios in which altruistic AVs can learn cooperative policies that are robust to diverse traffic scenarios and HV behaviors without compromising efficiency and safety, and present the results on transfer learning and domain adaptation in mixed-autonomy traffic.
\end{itemize}

The purpose of this chapter is to study the challenges of robust and safe AVs in mixed-autonomy traffic, especially in intrinsically competitive driving scenarios like those shown in Figure~\ref{fig:mainfigure}, in which coordination is essential for safety and efficiency. The intention is to utilize the autonomous driving challenge as a case study to examine the use of social theories from psychology literature in the MARL domain. To apply these theories to real-world roads, more study is required. Nonetheless, the research on altruistic AVs that are robust, safe, and capable of learning to influence HVs in desirable ways, without the limitations of current solutions are promising.

%


\section{Related Work}
\label{sec:relatedworks}
\subsection{Multi-agent Reinforcement Learning}
The intrinsic non-stationarity of the environment is a key problem for MARL. To address those limitations a MARL derivation of importance sampling is proposed and used to remove the outdated samples from the replay buffer~\cite{foerster2017learning}. In~\cite{xie2020learning} is presented another solution to address this issue by including latent representations of partner strategies allowing partner modeling and more scalable MARL.

To mitigate the problem of credit assignment in multi-agent systems,~\cite{foerster2018counterfactual} proposed the \textit{counterfactual multi-agent (COMA)} algorithm, which employs a centralized critic and decentralized actors. In ~\cite{egorov2016multi} is proposed a deep RL algorithm with full environment observability and a centralized controller to govern the joint-actions of all the agents. Other current research on mixed-autonomy focuses on addressing cooperative and competitive challenges by assuming the nature of interactions between autonomous agents~\cite{omidshafiei2017deep}. In~\cite{lowe2017multi} a variation of an actor-critic approach with a \textit{centralized q-function} is proposed. The algorithm has access to local observations and the actions of all agents. In our work, in contrast, we consider a decentralized controller with partial observability, and train altruistic agents that optimize for a social utility.

\subsection{Driver Behavior and Social Coordination}
Existing works on driver behavior and social navigation approach agents coordination by either modeling driver behaviors~\cite{brown2020taxonomy,ivanovicgenerative,mahjoub2019representing} or simplifying and making assumptions about the nature of agent interactions~\cite{lauer2000algorithm,omidshafiei2017deep}. In~\cite{toghi2020maneuver} is presented a maneuver-based dataset and a model for classifying driving maneuvers is proposed. 
Other works on driver behavior modeling consider graph theory~\cite{chandra2020cmetric}, data mining~\cite{constantinescu2010driving}, driver attributes~\cite{beck2014distress} or game theory~\cite{schwarting2019social}. In~\cite{ivanovicgenerative} is proposed a method for modeling and forecasting human behavior in circumstances that involve multi-human interactions in highly multi-modal situations.

Current research in social navigation has demonstrated the importance of AVs as social actors and the advantages of coordination between AVs and HVs~\cite{pokle2019deep}. Human driving patterns are learned from demonstration using inverse RL in~\cite{kuderer2015learning} and~\cite{sadigh2018planning}.  Similarly, in ~\cite{hadfield2016cooperative} is presented a centralized game-theory model for cooperative inverse RL. The authors in~\cite{trautman2010unfreezing} and~\cite{nikolaidis2015efficient} proposed a shared reward function to enable cooperative trajectory planning for robots and humans. Sadigh~\etal presents a strategy based on imitation learning to learn a reward function for human drivers, demonstrating how AVs can influence human actors~\cite{sadigh2016planning}. The importance of coordination and the advantage of using AVs to guide the traffic has been also investigated at the traffic level. Wu~\etal~\cite{wu2018stabilizing} analyzes the capability of AVs to stabilize a system of HVs and presents the conditions in which when concurrently enforcing safety constraints on the AVs while stabilizing traffic improves traffic performance. Similar works have highlighted the potential of influencing HVs and how AVs can be used to stabilize and guide the traffic flow~\cite{wu2018stabilizing, lazar2019learning}. 
Recent works focus on optimizing traffic networks in mixed autonomy to reduce traffic congestion and improve safety. In~\cite{biyik2021incentivizing} is presented a model of vehicle flow and a model of how AV makes decisions among routes with various prices and latencies. The planner optimizes for a social objective and shows improvement in traffic efficiency. The vehicle routing problem is studied in~\cite{ li2021learning} that proposed an innovative learning-augmented local search system to mitigate the problem by using a Transformer architecture. Cameron~\etal explores how humans can supervise agents in order to attain an acceptable degree of safety~\cite{hickert2021cooperation}. In contrast to previous works, we do not rely on human cooperation and our AVs learn cooperative behaviors directly from experience, our focus is on the emerging altruistic behavior that allows agents to coordinate and optimize for a social utility.

\subsection{Safe and Robust Driving}

Safety is critical for AVs~\cite{Safety_ECC}, and it is especially important for AVs that have been trained via RL. We must prioritize safety; because coordination is frequently associated with risk. In cooperative driving, there are often safe actions that have low rewards and riskier actions with higher rewards~\cite{wang2021emergent}; however, the risky action increases the likelihood of crashes when cooperation fails. Especially, AVs utilizing trained RL algorithms, may not always operate safely since the trained models may pick dangerous actions~\cite{li2018safe}. Several attempts in this direction use pure reward shaping to avoid collisions. While this is a frequent technique in RL, safety is not implicitly emphasized, and AVs implementing such RL methods may not behave properly in some cases due to function approximation.

To overcome this problem, the concept of safe RL is proposed in~\cite{li2018safe}, which aims to increase safety in unobserved driving conditions when the RL algorithm performs dangerously. ~\cite{wang2019lane} proposes a rule-based decision-making system that evaluates the controller's decisions and substitutes collision-causing actions. A short-horizon safety supervisor is included in Nageshrao~\etal~\cite{nageshrao2019autonomous}  to replace unsafe actions with safer ones. A Q-masking strategy is presented in~\cite{mohammadhasani2021reinforcement} to prevent collisions by deleting actions that might lead to a crash. Chen~\etal proposes a novel priority-based safety supervisor that reduces collisions considerably~\cite{chen2021deep}.

We leverage these approaches in this work using a decentralized reward function, local actions, and assuming partial observability, to increase the altruistic agents' safety while also being adaptable to varied driver behaviors and circumstances.
As shown in Figure~\ref{fig:mainfigure}, we analyze a particular situation in which AVs and HVs with various characteristics coexist. The picture depicts two frequent traffic situations in which vehicles must either merge into a lane effectively or depart the highway without colliding with other vehicles. In an ideal cooperative context, vehicles should proactively decelerate or accelerate to provide enough room for vehicles to safely exit/merge and prevent stalemate situations, while also being resilient to various conditions and behaviors and assuring safety in decision-making.


\section{Preliminaries and Formalism}
We study safety and robustness in the maneuver-level decision-making problem for AVs to see what kinds of behaviors might lead to socially desirable results. We're interested in the question of how AVs can be trained from scratch to drive safely and reliably, while also taking into account the social aspects of their mission, i.e., optimizing for a social utility that takes into account the interests of other vehicles in the vicinity. Social awareness and coordination are essential to improve safety and reliability on the roads. In this work, we explore that insight. Thus, we continue this section by providing a quantitative description of an agent's level of altruism and formally defined our problem.

It is possible to define the MARL problem as a centralized or decentralized problem. It's simple to create a centralized controller that provides a central joint reward and joint action. However, in the real world, such assumptions are unfeasible. In this chapter, we focus on a decentralized controller with partial observability and formulate the problem as a partially observable stochastic game (POSG) defined by $\langle \mathcal{I}, \mathcal{S}, P, \gamma, \{ \mathcal{A}_i \}_{i\in\{1,...,N\}}, \{ \mathcal{O}_i \}_{i\in\{1,...,N\}}, \{ R_i \}_{i\in\{1,...,N\}}  \rangle$ where
\begin{itemize}
\item $\mathcal{I}$: a finite set of agents $N \geq 2$. 
\item $\mathcal{S}$ : a set of possible states that contains all configurations that $N$ AVs can take (probably infinite).
\item $P$: a state transition probability function from state $s \in \mathcal{S}$ to state $ s' \in \mathcal{S}$, $P(S=s'|S=s,A=a)$.
\item $\gamma$: a discount factor, $\gamma \in [0, 1]$. 
\item $\mathcal{A}_i$: a set of possible actions for agent $i$.
\item $\mathcal{O}_i$: a set of observations for agent $i$.

\item $R_i$: a reward function for the $i^{th}$ agent, $R_i(s,a)$. 

\end{itemize}

 At a given time $t$ the agent senses the environment and receives a local observation $o_i: \mathcal{S} \rightarrow \mathcal{O}_i$, based on the observation $o_i$ and its stochastic policy $\pi_i: \mathcal{O}_i \times \mathcal{A}_i \rightarrow [0, 1]$, the agent takes an action within the action-space $a_i \in \mathcal{A}_i$. Consequently, the agent transits to the next state $s'$ which is determined based on the state transition probability function $P(s'|s, a): \mathcal{S} \times \mathcal{A}_1 \times ... \times \mathcal{A}_N \rightarrow \mathcal{S} $ and receives a decentralized reward $r_i: \mathcal{S} \times \mathcal{A}_i \rightarrow \mathbb{R}$. The goal of each agent $i$ is to optimally solve the POSG by deriving a probability distribution over actions in $\mathcal{A}$ at a given state, that maximizes its cumulative discounted sum of future rewards over an infinite time horizon and find the corresponding optimal policy $\pi^*: \mathcal{S} \rightarrow \mathcal{A}$.

An optimal policy maximizes the action-value function, i.e., 
\begin{equation}
\label{equ:policy}
\pi^*(s) = \arg\max_a Q^* (s,a) 
\end{equation}

where, 

\begin{equation}
\label{equ:q}
Q^\pi(s,a) \coloneqq \mathbb{E}_{\pi} [\sum_{k=0}^\infty \gamma^k R_k(s,a) |s_0=s, a_0=a].
\end{equation}

 The optimal action-value function is determined by solving the Bellman equation,
\begin{equation}
\label{equ:bellmanequ}
Q^*(s,a) = \mathbb{E} \left[ R(s,a) + \gamma \max_{a'}  Q^*(s',a') |s_0=s, a_0=a \right]
\end{equation}




\subsection{Double Deep Q-Network}
Deep Q-network (DQN) has been widely used in RL problems. DQN uses a deep neural network (NN) with weights $\textbf{w}$ as the function approximator to estimate the state-action value function, i.e., $\Tilde{Q}(.;\textbf{w}) \cong Q(.)$. DDQN improves DQN by decomposing the max operation in the target into action selection and action evaluation, mitigating the over-estimation problem.
The idea is to periodically sample data from a buffer and compute an estimate of the Bellman error or loss function, written as
\begin{equation}
\label{equ:loss2}
\mathcal{L}(\textbf{w}) = \mathbb{E}_{s,a,r,s' \sim \mathcal{RM}}[( Target - \Tilde{Q}(s,a;\textbf{w}))^2]
\end{equation}
\vspace{-5pt}
\begin{equation}
\label{equ:DDQNtarget}
Target = R(s,a) + \gamma \Tilde{Q}(s',\underset{a'}{\arg\max} \Tilde{Q}(s',a';\textbf{w});\hat{\textbf{w}}))
\end{equation}

The DDQN algorithm then performs mini-batch gradient descent steps as $\textbf{w}_{i+1} = \textbf{w}_i - \alpha_i \hat{\nabla}_\textbf{w} \mathcal{L}(\textbf{w})$, on the loss $\mathcal{L}$ to learn the approximation of the value function ($\Tilde{Q}(.)$). The $\hat{\nabla}_\textbf{w}$ operator denotes an estimate of the gradient at $\textbf{w}_k$,  $\textbf{w}$ are the weights of the online network and $\hat{\textbf{w}}$ are the weights of the target network which are updated at a lower frequency ($Target_{update}$) to stabilize training. The experience replay buffer ($RM$) is used to generate training samples $(s, a, r, s')$, which are randomly drawn to  protect from correlated observations and non-stationary data distribution. 



\subsection{Driving Scenarios}
Our objective is to investigate driving scenarios in which the lack of AV coordination hinders safety and efficiency. We also study adaptability among scenarios and driver behaviors. For this, we design a set of scenarios $\mathcal{F}$ with highway exiting and merging ramps as the main scenarios, as shown in Figure~\ref{fig:mainfigure}, where a mission vehicle (in our case an exiting/merging vehicle) attempts to accomplish its task in a mixed-traffic environment.

The exiting and merging scenarios are designed in such a way that coordination is necessary for safety. AVs must coordinate, and neither can achieve a safe and smooth traffic flow on its own, i.e., exiting/merging will not be feasible without the coordination of the other AVs. To facilitate safe exiting/merging while also responding to varied traffic scenarios, altruistic AVs must learn to account for the interests of all vehicles, coordinate, make compromises, and influence human behavior. 
In Figure~\ref{fig:mainfigure}, for example, the AV1 has to compromise its own utility and reduce speed to guide the traffic of aggressive HVs, creating space for the exiting/merging vehicle, while the other AVs have to increase speed to create room for the mission vehicle. The exiting and merging scenarios are defined as $f_e , f_m \in \mathcal{F} $ correspondingly. We particularly chose those scenarios as a case of study because of their intrinsic similarity and the need for coordination, as the exiting/merging vehicle's utility contrast with that of the HV highway vehicles.

\subsection{Social Value Orientation for AVs}
In this section, we introduce \textit{Social Value Orientation (SVO)} to formally investigate the social conflicts between humans and agents in diverse environments. It is critical to quantify an individual's social preference to understand whether they would cooperate or not in a particular scenario, such as opening a gap in our highway merging example. For that purpose, SVO is a commonly used concept in the social psychology literature that has lately been applied in robotics research~\cite{schwarting2019social}. In our context, SVO defines the degree of an agent's egoism or altruism toward others. Based on the value placed on the utility of others, an HV or an AV's behavior can range from egoistic to completely altruistic. We rely on AVs to guide traffic toward more socially advantageous outcomes since the SVO of HV is unknown. In formal terms, an AV's SVO angle $\phi$ determines how the AV balances its own reward against that of others~\cite{toghi2021cooperative, toghi2021social,le2022cooperative}. In terms of rewards, an AV's total reward $R_i$ is defined as:
\begin{equation}
\label{equ:svodefinition}
R_i = r_i \cos \phi_i + r^-_i \sin \phi_i
\end{equation}
where $r_i$ is the agent's individual utility, $r^-_i$ is the total utility of other agents from the perspective of the $i$th agent which in general is a function $f(.)$ of their individual utilities,
\begin{equation}
\label{equ:othersutility}
r^-_i = f(r_j), \quad \text{where } j \neq i 
\end{equation}
The SVO angle can varied from $\phi=0$ (entirely selfish) to $\phi=\pi/2$ (entirely altruistic). Nonetheless, none of the limits are optimal, and a point in the middle, known as the optimal SVO angle~$\phi^*$ gives the most socially favorable outcome. SVO allows us to understand the behaviors that make possible the socially desirable outcomes in Figure~\ref{fig:mainfigure}. 

\begin{figure}[t]
\centering
\includegraphics[width=.8\textwidth]{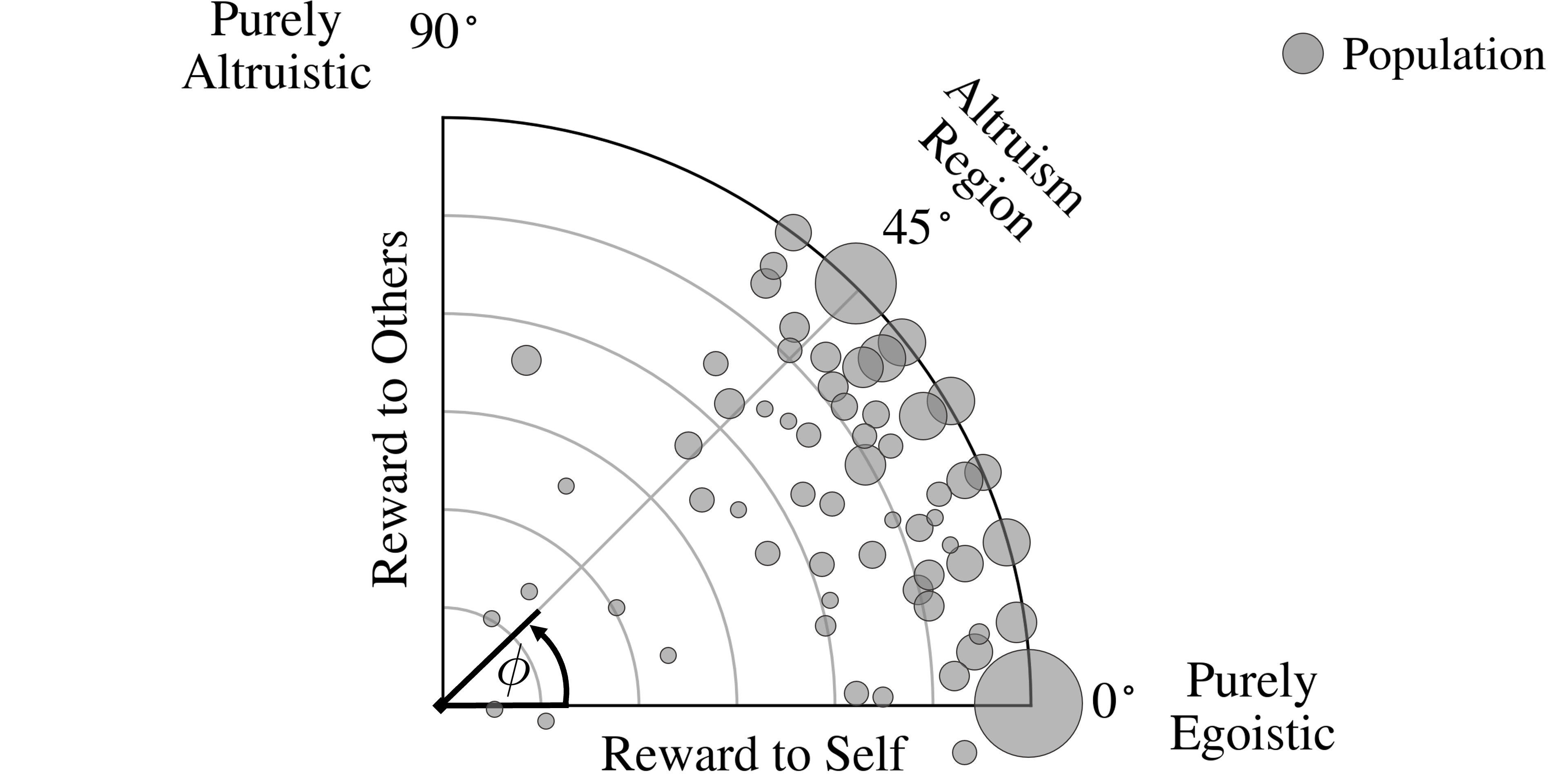}
\caption{\small{The SVO angle $\phi$ quantifies the level of altruism of an agent. In the figure the diameter of the circles, represents the size of the human population that holds the associated SVO~\cite{garapin2015does}.}}
\label{fig:svoring}
\end{figure}

Autonomous agents must be aware of human drivers' social preferences as well as their desire to collaborate. Humans, on the other hand, are known to be diverse in SVO, and so their preferences are uncertain~\cite{murphy2015social}. Figure~\ref{fig:svoring} depicts a range of altruism across individuals with varying SVO. As a result of the wide range of altruistic behavior seen in humans, is not safe to rely on humans to guide the traffic, instead, we should rely on AV to guide the traffic toward more socially advantageous goals. Therefore, our objective is that the AVs learn to create alliances and influence HV behavior to improve the global utility of the group.

%

\subsection{Autonomous Vehicles as Social Actors}
AVs in a mixed environment will be social actors in the traffic road that will react to HVs and influence and adapt to their behaviors. The traffic environment is rich with habits and expectations, that determine driver behaviors. The vehicle's interaction will be influenced by the way AV decisions are perceived~\cite{socialmuller,schwarting2019social}.
For instance, some human drivers may be grateful if the AV stops for them but frustrated if it does not perform as expected. Also, they might behave aggressively if they're stuck behind an overly cautious AV, which reduces speed constantly.  Another example is the case that when crossing a street while a vehicle is waiting, pedestrians move faster (a gesture of respect for the driver). On the other hand, will pedestrians speed up for an AV, or will they behave differently?
If an AV is understood as a social actor, the HVs will learn the individual and social traits of AVs and behave accordingly in mutual interactions. This would fit with current preconceptions that make assumptions about drivers based on the brand and type of vehicle they drive. Current AVs' driving is as conservative as possible to ensure safety. They will slow down in front of a crossing because they believe the other vehicle will want to go first, even though this is against the law. They wait for pedestrians when in doubt. It's not difficult to see how other agents and HVs could take advantage of and exploit these over-conservative behaviors. As AVs are going to be social actors in mixed autonomy traffic, the safety and reliability of AVs will be coupled with their social awareness and their ability to engage in complex social interactions. We consider risk awareness and social behavior as fundamental traits for decision-making.

Failure to identify social cues of selfishness or collaboration by an AV has ramifications for the general flow of the traffic network, as well as the safety of traffic participants. Current AVs ignore social signs and driver personality in favor of explicit communication or driver modeling. Because these methods can't handle complicated interactions, they tend to be conservative, restricting autonomy solutions to simple road interactions~\cite{socialmuller,schwarting2019social}.
The ability of AVs to drive in a socially obedient manner is critical for the safety of passengers and other vehicles because predictable behavior allows humans to comprehend and respond appropriately to the AV's actions.

\subsection{Driving Behaviors}
The problem of simulating varied behaviors may be defined as determining the appropriate range of parameters to produce heterogeneous behaviors within the simulator. Some works in social traffic psychology show that driving behavior falls between conservative and aggressive. Nevertheless, the specific definition is still under discussion and fluctuates across works~\cite{sagberg2015review}.
The phrase "aggressive driving" refers to a wide range of unsafe driving practices, including running red lights and speeding. The root of aggressive driving has a variety of factors that aren't necessarily clear. Some are caused by hazardous road conditions, while others are caused by personal characteristics or mental states.
Moreover, there is a correlation between aggressiveness and egoism, as egoistic drivers are less likely to yield and have a tendency to over-speeding and engage in unsafe actions. While there is a correlation between these concepts~\cite{harris2014prosocial,vallieres2014intentionality}, we distinguish aggressiveness from egoism in this study by describing individual traits and social preferences.

In this work, we discriminate between individual traits and social preferences because they result in different behaviors. We define altruism and egoism as social preferences; in that sense, an egoistic driver is a selfish driver who accounts for his personal utility irrespective of his aggression. We define conservatism and aggressiveness as individual traits, and we describe an aggressive driver as someone whose actions result in aggressive behavior. Individual traits such as aggressiveness are characterized by the outcomes of their actions, but social preferences such as egoism are distinguished by their social objectives and purposes. In this direction, an egoistic driver is a self-centered driver who lacks social motive, a driver who believes he controls the road and disregards the other drivers. Egoist drivers frequently engage in violent actions, and while ego defensiveness is not the primary source of aggression, it is a major contributor to aggressive driving~\cite{harris2014prosocial,vallieres2014intentionality}. Despite their similarities, the two groups have different origins and result in different behaviors. A driver, for example, could be egoistic and conservative. We may envision a driver who drives cautiously to protect himself (selfish motivation/preference) and, as a result, is conservative in his behavior (outcome of his actions).

Properly, we described social preferences (altruism or egoism) by the AV's SVO angular phase $\phi$; and individual traits (conservativeness and aggressiveness) by the HV driver model parameters ($\mathcal{P}$) as described in section~\ref{sec:humandrivermodel}. Based on the values of these parameters, a driver will behave conservative or aggressive. In the simulations, the AVs have no access to HVs' SVO, we consider the SVO of HVs to be undetermined as they cannot communicate that directly.
Finally, we define a set of behaviors $\mathcal{B}$, i.e, aggressive, moderate and conservative, $b_a,b_m,b_c \in \mathcal{B}$ based on the parameters ($\mathcal{P}$) obtained in section~\ref{sec:humandrivermodel}.

 
\section{Problem Formulation}
\label{sec:problem_formulation}
We investigate the safety and robustness of the scenarios described in Figure~\ref{fig:mainfigure}, an exiting/merging vehicle, which can be either HV or AV. This configuration contains a group of AVs that hold the same SVO, as well as a group of HVs which are heterogeneous in their SVO, making it unclear whether they are allies or opponents. Formally, the road is shared by a set of HVs $h_k \in \mathcal{H}$, with an undetermined SVO $\phi_k$ and heterogeneous behaviors $b_k \in \mathcal{B}$; a set of AVs $i_i \in \mathcal{I}$, that are connected together using V2V communication, controlled by a decentralized policy and sharing the same SVO, and a \emph{mission vehicle}, $M \in \mathcal{I} \cup \mathcal{H}$ that is aiming to accomplish its mission (highway exiting/merging) and it can be either AV or HV. We focus on the multi-agent maneuver-level decision-making problem for AVs in mixed-autonomy environments and study the following problems: how AVs can learn in a mixed-autonomy environment optimal cooperative policies $\pi^*(s)$ that are robust to different scenarios $f \in \mathcal{F}$ and behaviors $b \in \mathcal{B}$ while ensuring safety on the decision-making, and how sensitive is the performance of the altruistic AVs to the HVs' behaviors. 

As AVs are connected, we assume that they receive an accurate local observation of the environment $\Tilde{\textbf{o}}_{i} \in \widetilde{\mathcal{O}}_i$, sensing all the vehicles within their perception range, i.e, a subgroup of HVs $\widetilde{\mathcal{H}} \subset \mathcal{H}$ and a subgroup of AVs $\widetilde{\mathcal{I}} \subset \mathcal{I}$. Nevertheless, AVs are unable to share their actions or rewards, and they take individual actions from a set of high-level actions $a_i \in \mathcal{A}_i (|\mathcal{A}_i|=5)$. 
The goal of this work is to train social-aware AVs that learn how to drive in a mixed-autonomy scenario in a robust, efficient, and safe manner. We are interested in how to obtain a utility function that enables AVs to handle competitive driving scenarios (such as those in Figure~\ref{fig:mainfigure}) and leads them into socially-desirable decisions that improve traffic efficiency, safety, and robustness.


\section{Safe and Robust Social Driving}
\label{sec:solution}
In this section, we present the safe and robust MARL approach. Our approach uses a general decentralized reward function that optimizes for social utility and induces altruism in the AVs; the general reward function accounts for any anticipated vehicle's mission, allowing it to be applied to a variety of environments; and collisions are reduced by the safety prioritizer.
What we define as "driving" is the outcome of decades of human learning from experience. Consequently, we take the same approach and train AVs that learn from experience and define the optimization problem as the eventual desirable social outcome with adaptability, expecting AVs to learn how to drive safely during the process.
We carefully design a decentralized general reward function, a suitable architecture, and a safety prioritizer to promote the desired safe altruistic behavior in AVs' decision-making process. The overview of our approach as presented in Figure~\ref{fig:cover} and Figure~\ref{fig:mainfigure} helps us to create intuition on these points, by introducing driving scenarios in which altruistic AVs lead to socially advantageous results while adapting to different traffic scenarios. 

\textbf{Action Space:}
The goal of this research is to look at inter-agent and agent-human interactions, as well as behavioral elements of mixed-autonomy driving. Thus, we choose a more abstract level and define the action-space as a set of discrete meta-actions $a_i \in  \mathcal{A}_i$. In particular, we select a set of five high-level actions $a_i$ as,
\begin{equation}
\label{equ:action_space}
    a_i \in \mathcal{A}_i =
    \begin{bmatrix}
        \texttt{Lane Left}\\
        \texttt{Idle}\\
        \texttt{Lane Right}\\
        \texttt{Accelerate}\\
        \texttt{Decelerate}
    \end{bmatrix}
\end{equation}
%
These meta-actions are then converted into trajectories and low-level control signals, which ultimately control the vehicle's movement.

\begin{figure}[t]
  \centering
  \includegraphics[width=.9\textwidth]{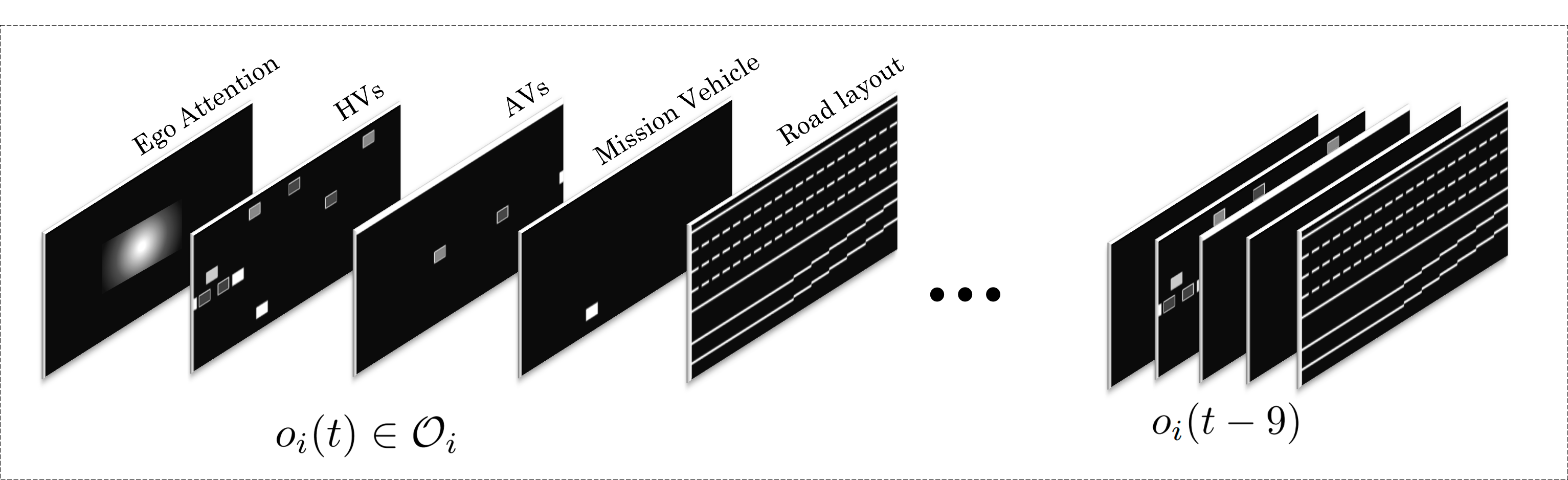}
\caption{\small{Multi-channel VelocityMap state representation embeds the speed of the vehicle in pixel values.}}
  \label{fig:heatmap}
\end{figure}

\textbf{Observation Space:}
We use a \textit{multi-channel VelocityMap} observation ($o_i$) that embeds the relative speed of the vehicle with respect to the ego vehicle in pixel values~\cite{toghi2021social}. We represent the information in multiple semantic channels that embed: 1) an attention map to highlight the position of the ego vehicle, 2) the HVs, 3) the AVs, 4) the mission vehicle, and 5) the road layout. Figure~\ref{fig:heatmap} illustrates an example of this multi-channel representation.
In order to map the relative speed of the vehicles into pixels, we use a clipped logarithmic function, which improves dynamic range and yields better results than a linear map, i.e.,
\begin{equation} \label{equ:logmapping}
Z_j = 1 - \beta \log (\alpha |v_j^{(l)}|) \mathds{1}(|v_j^{(l)}|-v_0) 
\end{equation}
where $Z_j$ is the pixel value of the $j$th vehicle in the state representation, $v^{(l)}$ is its relative Frenet longitudinal speed from the $k$th vehicle's point-of-view, i.e., $\dot{l_j}-\dot{l_k}$, $v_0$ is speed threshold, $\alpha$ and $\beta$ are dimensionless coefficients, and $\mathds{1}(.)$ is the Heaviside step function. Such non-linear mapping gives more importance to neighboring vehicles with smaller $|v^{(l)}|$ and almost disregards the ones that are moving either much faster or much slower than the ego vehicle.
As temporal information is necessary for safe decision-making, we use a history of successive VelocityMaps observations to create the input state to the Q-network.

%
%

%
\subsection{Distinguishing Sympathy from Cooperation}
\label{sec:sympcoop}
In our mixed-autonomy problem, we divide inter-agent relations into interactions between autonomous agents (AV-AV interactions) and interactions between autonomous agents and human drivers (HV-AV interactions). By decoupling the two, we can analyze the interactions between human drivers with unclear SVO and our autonomous agents in a methodical way. In that sense, we define \textit{sympathy} as the autonomous agent's altruism toward a human, and \textit{cooperation} as the altruistic behavior among autonomous agents. The fact that the components of altruism differ in nature is our reasoning for separating them. Sympathy, for example, may not be reciprocated since agents differ in their SVO, whereas cooperation among autonomous entities is fundamentally homogeneous if they share the same SVO.
%
Following this concept, we can rewrite the AV reward in Eq.~\eqref{equ:svodefinition} as,
\begin{equation} \label{equ:sympcoop}
\begin{aligned}
R_i  ={} & r_i \cos \phi_i +(\sin \theta_i R_i^{\mathrm{AV}} + \cos \theta_i R_i^{\mathrm{HV}}) \sin \phi_i \\
 = &  \underbrace{r_i \cos \phi_i}_{\mathrm{egoistic \,\, term}} + \\
& \underbrace{\sin \theta_i \sin \phi_i R_i^{\mathrm{AV}}}_{\mathrm{cooperation \,\, term}} + \underbrace{\cos \theta_i \sin \phi_i R_i^{\mathrm{HV}}}_{\mathrm{sympathy \,\, term}}
\end{aligned}
\end{equation}
where $\theta$ is the sympathy angular phase determining the cooperation-to-sympathy ratio. Parameters $R_i^{\mathrm{AV}}$ and $R_i^{\mathrm{HV}}$ denote the total utility of other AVs and HVs, respectively, as perceived from the $i$th agent's perspective. We expand on this topic in Section~\ref{sec:reward} where we introduce the distributed reward structure.
%

\subsection{Decentralized Social Reward}
\label{sec:reward}
The AVs are trained using the partial local observations and the decentralized reward function, and we anticipate them to learn how to drive in a variety of settings while taking into consideration the individual diver's missions. As a result, we create a well-engineered general reward function that considers social utility, traffic metrics, and individual diver's missions.
Following the definition of sympathy and cooperation in equation~\eqref{equ:sympcoop} we decompose the decentralized reward received by agent $I_i \in \mathcal{I}$ as,
\begin{equation} \label{equ:decentralizedreward}
\begin{aligned}
R_i(s, a) ={} & R^{\mathrm{ego}}+R^{\mathrm{social}} 
\\R^{\mathrm{ego}} = {} & \cos \phi_i r_i(s, a) 
\\R^{\mathrm{social}}  = {} &  R^{\mathrm{coop}} + R^{\mathrm{symp}} 
\\R^{\mathrm{coop}} = {} & \sin \theta_i \sin \phi_i \Big[ \sum_j r^{\mathrm{AV}}_{i, j} (s, a)+ \sum_j r_{i,j}^M (s, a)\Big]
\\R^{\mathrm{symp}} = {} &  \cos \theta_i \sin \phi_i \Big[ \sum_k r^{\mathrm{HV}}_{i, k} (s, a) + \sum_k r_{i,k}^M (s, a) \Big]\\
\end{aligned}
\end{equation}

in which $R^{\mathrm{ego}}$, $R^{\mathrm{social}}$ represents the egoistic and social reward, $i \in \mathcal{I} $, $j \in (\widetilde{\mathcal{I}} \setminus \{I_i\})$, $k \in \widetilde{\mathcal{H}}$. The term $r_i$ represents the ego vehicle's reward obtained from traffic metrics and the angle $\phi$ allows to adjust the level of egoism or altruism.
$R^{\mathrm{coop}}$ is the cooperation term (the altruistic behavior among AVs, i.e, AV's altruism toward others AVs) and $R^{\mathrm{symp}}$ is the sympathy term (AV's altruism toward HVs).
The sympathy reward term, $r^{\mathrm{HV}}_{i, k}$ considers the individual reward of the HVs, while the cooperation reward term, $r^{\mathrm{AV}}_{i, j}$ considers the individual reward of the other AVs, and are defined as
\begin{equation} \label{equ:symreward}
r^{\mathrm{HV}}_{i, k} = \frac{\mathcal{W}_k}{d_{i,k}^\lambda} \sum_m \omega_m x_m \quad 
r^{\mathrm{AV}}_{i, j} = \frac{\mathcal{W}_j}{d_{i,j}^\lambda} \sum_m \omega_m x_m
\end{equation}

in which $d_{i,k}/d_{i,j}$ represents the distance between the agent and the corresponding HV/AV, $\lambda$ is a dimensionless coefficient, $\mathcal{W}_k$ is a weight value for individual vehicle's importance, $m$ is the set of traffic metrics that have been considered in the vehicle's utilities (speed, crashes, etc.), in which $x_m$ is the $m$ metric normalized value and $w_m$ is the weight associated to that metric. The term $r^{\mathrm{M}}$ accounts for the reward of the vehicle's mission. A mission is defined as any desired specific outcome for a particular vehicle, as merging, exiting, etc. 
%
\begin{equation}
\label{equ:missionreward}
r^{\mathrm{M}}_{i,j} = 
\begin{cases}
\frac{w_j}{(d_{i,j})^\mu}, & \mathrm{if} g(j) \\
0, & \mathrm{o.w.}
\end{cases}
\quad
r^{\mathrm{M}}_{i,k} = 
\begin{cases}
\frac{w_k}{(d_{i,k})^\mu}, & \mathrm{if} g(k) \\
0, & \mathrm{o.w.}
\end{cases}
\end{equation}

The function $g(v)$ is an independent function to evaluate the mission; $g(v)$ returns true if the vehicle $v$ has a mission defined and the mission has been accomplished in the recent time window. $\mu$ is a dimensionless coefficient,
$w_j/w_k$ are weights for an individual vehicle's mission (importance of the mission).
This allows defining a general reward independent of the driving scenario and mission goals for different vehicles. In the experiments, a \textbf{HV} can be assigned a merging mission or a highway exiting mission, as referred to in Figure~\ref{fig:mainfigure}.


\subsection{Deep MARL architecture for Social driving}
\begin{figure}[t]
  \centering
  \includegraphics[width=.9\textwidth]{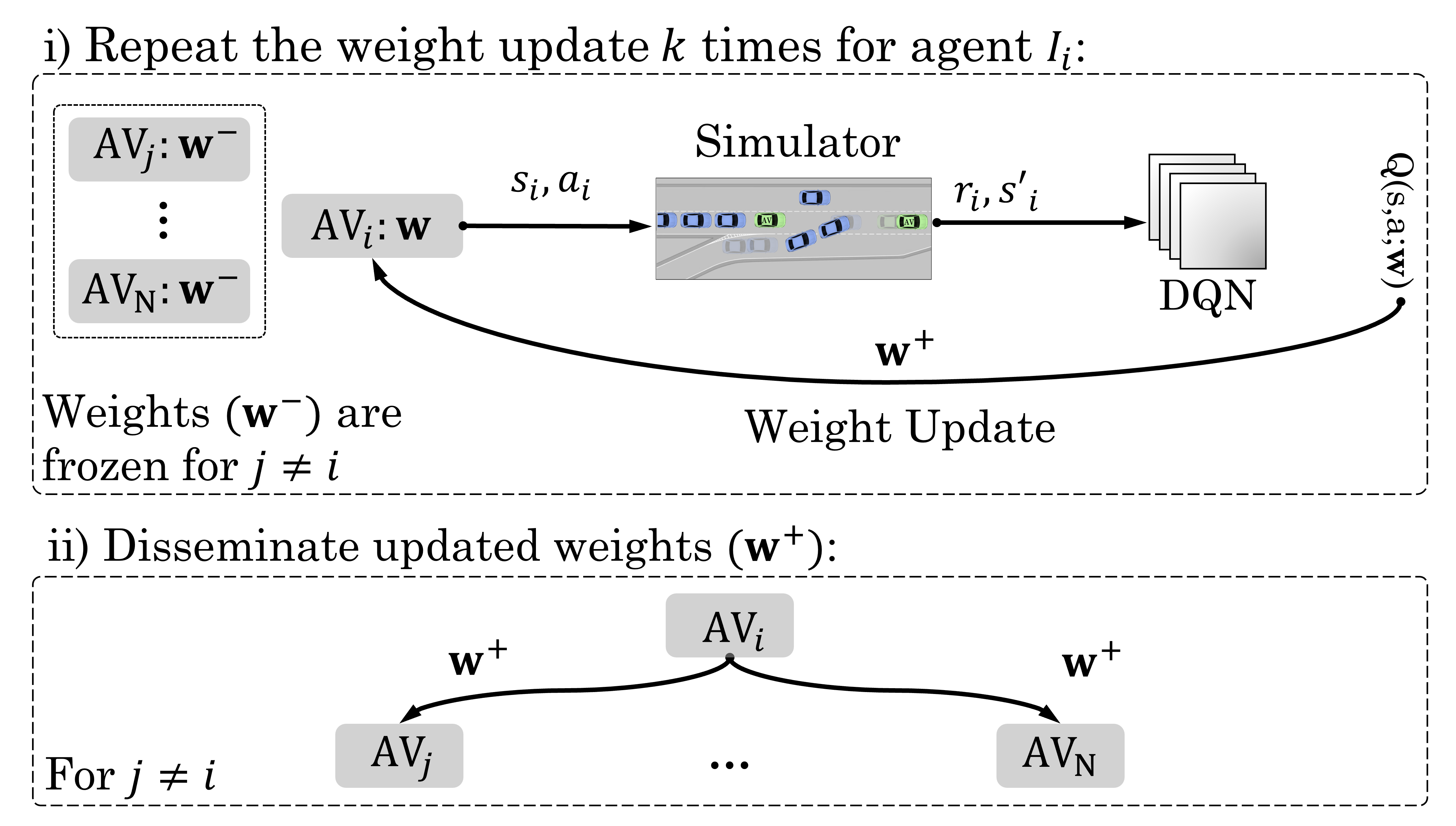}
  \caption{\small{The multi-agent training and policy dissemination process.}}
  \label{fig:coordinatedPOSG}
\end{figure}

\begin{figure}[t]
  \centering
  \includegraphics[width=.9\textwidth]{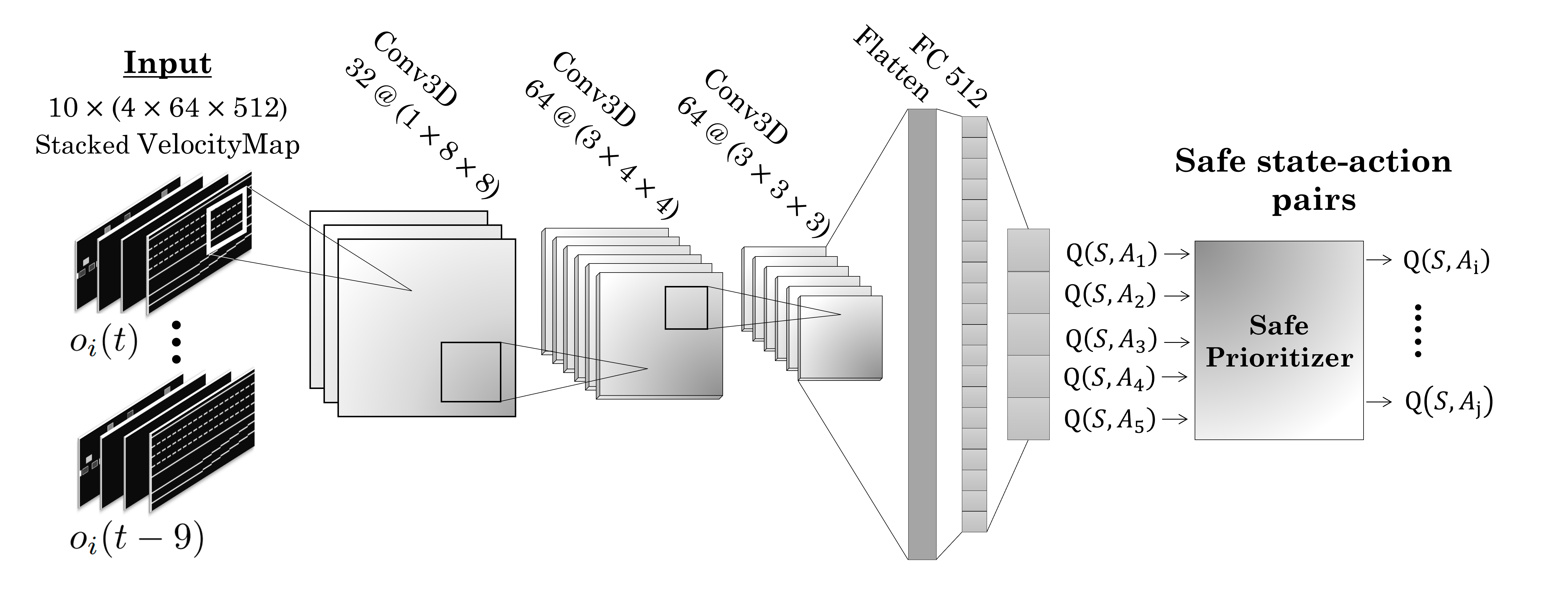}
  \caption{\small{Deep MARL architecture with the safety prioritizer.}}
  \label{fig:architecture}
\end{figure}

As shown in Figure~\ref{fig:architecture}, we leverage a 3D Convolutional Neural Network (CNN) with a safety prioritizer for our MARL architecture. To account for the temporal information, the 3D CNN operates as a feature extractor and leverages a history of VelocityMap observations. The network receives a stack of 10 VelocityMap observations, i.e., a $10 \times (4 \times 512\times 64)$ tensor that captures the latest 10 time-steps episodes.
To mitigate the non-stationarity issue in MARL, agents are trained in a semi-sequential manner, as illustrated in Figure~\ref{fig:coordinatedPOSG}. The agents are trained independently for $N_{iterations}$ iterations while freezing the policies of the remaining AVs, $\textbf{w}^-$. Subsequently, the other agents' policies are updated with the new policy, $\textbf{w}^+$.

To improve safety we train our agents using a safety prioritizer that, in the cases where the action selected by the agent policy is unsafe, selects a safe action and stores the unsafe action ($a_t$) and the related state in the $RM$ with a suitable penalty on the reward ($r_{unsafe}$) for the unsafe state-action pair. The safety prioritizer reduces episode resets due to imminent collisions improving sample efficiency. The unsafe state-action pairs are not removed so the agent can also learn from unsafe experiences. The experience $(\psi(s_{t}), a_t , r_{unsafe} , \emptyset$) is stored in $RM$ with a terminal next state $\emptyset $, the target for this unsafe pair $ (s_t,a_t)$ is $Target(s_t, a_t)^{DDQN} = r_{unsafe}$. 
The details of the safety prioritizer are given in the next section~\ref{sec:safeprioritizer}.

The proposed deep MARL architecture is described in \textbf{Algorithm}~\ref{alg:Robust_MARL_algorithm}. As part of the implementation, we start the learning process after the replay buffer has been filled with a sufficient number of sample simulations. Furthermore, we update the experience replay buffer to adjust for the extremely skewed training data~\cite{toghi2021social}. Balancing skewed data is a frequent practice in machine learning, and it was effective in our MARL problem.

\begin{algorithm}[t]
    \caption{Safety Prioritized Multi-agent DDQN} 
    \label{alg:Robust_MARL_algorithm}
    \begin{algorithmic}
         \STATE Initialize \textit{experience replay buffer} $RM$.
         \STATE Initialize $\Tilde{Q}(.;\textbf{w}^-)$ with random weights $\textbf{w}^-=\textbf{w}_{ini}$
         \STATE Initialize target network $\Tilde{Q}(.;\hat{\textbf{w}})$ with weights $\hat{\textbf{w}}=\textbf{w}^-$
          \STATE Pre-store experience of first's 50 episodes in $RM$
         \FOR{$\mathrm{e}=50$ to $N_{\mathrm{episode}}$} 
         \STATE Initialize $s_1 =\{ \Tilde{\textbf{o}}_{1}  \}$  and compute $\psi_1 = \psi(s_1)$
              \FOR{t = 1 to T }
                \FOR{$I_i$ in $\mathcal{I}$}
                    \STATE Freeze $\textbf{w}^-$ for all $I_j$, $j \neq i$
                     \FOR{$m=1$ to $N_{iterations}$}
                         \STATE With probability $\epsilon$ select a random action $a_t$ ,
                         \STATE otherwise select $a_t = \max_{a' \in A} Q(\psi(s_t),a';\textbf{w}^+)$
                         \IF {$a_t$ is unsafe (\textbf{Algorithm}~\ref{alg:safecheck})} 
                          \STATE Store $(\psi_t, a_t , r_{unsafe} , \emptyset$) in $RM$
                         \STATE $a_t$ = Compute a safe action (\textbf{Algorithm}~\ref{alg:safeaction}) 
                         \ENDIF
                          \STATE Execute safe action $a_t$ , and observe $r_t, \Tilde{\textbf{o}}_{t}$ 
                         \STATE Set $s_{t+1} =\{  s_{t},\Tilde{\textbf{o}}_{t+1} \}$ and $\psi_{t+1} = \psi(s_{t+1})$
                         \STATE Store experience $(\psi_t, a_t , r_t , \psi_{t+1}) $ in $RM$
                         \STATE Sample a mini-batch of size $M$ from $RM$
                        \STATE Compute $\mathcal{L}(\textbf{w}^+)$ 
                        \STATE Performs gradient descent 
                        \STATE $\textbf{w}^+_{k+1} \leftarrow \textbf{w}^+_k - \alpha \hat{\nabla}_\textbf{w} \mathcal{L}(\textbf{w}^+)$

              \ENDFOR
              \STATE $\textbf{w}^- = \textbf{w}^+$ for all $I_i \in \mathcal{I}$
            \ENDFOR
            \STATE Every $\quad Target_{update} \quad$ reset $\hat{\textbf{w}} \leftarrow \textbf{w}^-$
          \ENDFOR
        \ENDFOR
    \end{algorithmic}
\end{algorithm}


\subsection{Safety Prioritizer}
\label{sec:safeprioritizer}
We include a safety prioritizer to the MARL algorithm that penalizes and reduce imminent crashes. This helps the agent to increase sample efficiency during training and avoid collisions when in deployment. If the agent comes into an unexpected situation and decides to perform a risky action, that action will be prevented. The safety prioritizer enhances simulation results and is crucial in real-world scenarios. The safety prioritizer included~\textbf{Algorithm}~\ref{alg:safecheck} and \textbf{Algorithm}~\ref{alg:safeaction}.

\textbf{Algorithm}~\ref{alg:safecheck}: During action selection of the agent $I_i$, once an action $a_t$ is chosen, the safety prioritizer checks if the action is safe by computing a safety score for $N_{steps}$ of planning.
We utilize the time-to-collision ($ttc$) as a safety score. If $safety_{score}< safe_{th}$ the action is unsafe and we need to select a safe action. The selection of a safe action is presented in \textbf{Algorithm}~\ref{alg:safeaction}. 

\textbf{Algorithm}~\ref{alg:safeaction}: The safe action selection is different in training and testing. During training, to encourage exploration, we remove the unsafe actions and keep the random action selection following the current exploration policy on the remaining actions. During testing, we follow the greedy policy in the subset of safe actions $a_t = \max_{a' \in \widetilde{\mathcal{A}}_{safe} } Q(\psi(s_{t}),a';\textbf{w})$. It should be noted that the algorithm does not choose the safest of all possible actions, as that action may lead to particularly conservative behaviors that can compromise traffic efficiency; we instead remove the imminent unsafe actions and follow the priority given by the learned altruistic policy. If it happens that all possible actions are unsafe, we return the action $a_t \in \mathcal{A}$ with the highest safety score. In that way during training the constrained exploration will keep the agent from taking unsafe actions which will lead to efficient sampling and more stable learning; and during testing, the decision-making is based on the prosocial learned policy with minimum intervention from the safety prioritizer, achieving higher traveled distance while avoiding collisions. 

\begin{algorithm}[t]
    \caption{Safety score} 
    \label{alg:safecheck}
    \begin{algorithmic}
         \STATE Simulate $I_i$ taking the action $a_t$
         \FOR{$v$  in $(\widetilde{\mathcal{I}} \cup  \widetilde{\mathcal{V}})  \setminus \{I_i\}$ } 
         \STATE Compute safety score of $I_i$, $v$ for $N_{steps}$ planning 
         \IF{$safe_{score}$ $ <  safe_{th}$}
         \STATE Return unsafe
         \ENDIF
         
        \ENDFOR
         \STATE Return safe
    \end{algorithmic}
\end{algorithm}

\begin{algorithm}[t]
    \caption{Safe action} 
    \label{alg:safeaction}
    \begin{algorithmic}
        \STATE Initialize  $\widetilde{\mathcal{A}}$ = $\mathcal{A}$
        \WHILE{$\widetilde{\mathcal{A}}$ is not empty}
        \IF{during training}
        \STATE Select $a_t$ following the exploration policy on set $\widetilde{\mathcal{A}}$ 
        \ELSIF{during test}
            \STATE Select$a_t = \max_{a' \in \widetilde{\mathcal{A}}_{safe} } Q(\psi(s_{t}),a';\textbf{w})$ 
        \ENDIF
        \IF{$a_t$ is safe (Algorithm \ref{alg:safecheck})}
        \STATE Return $a_t$ 
        \ENDIF
     \ENDWHILE
     \STATE Return $a_t$ with highest safe score in $\mathcal{A}$
    \end{algorithmic}
\end{algorithm}

\subsection{Modeling Driver Behaviors} 
\label{sec:humandrivermodel}
We model the longitudinal movements of HVs using the \textit{Intelligent Driver Model} (IDM)~\cite{treiber2000congested}, while the lateral actions of HVs are based on the MOBIL model~\cite{kesting2007general}. The MOBIL model considers two main criteria,

\textbf{The safety criterion} ensures that after the lane change, the deceleration of the new follower $\mathrm{a}^{}_n$ in the target lane does not exceed a safe limit, i.e, $\mathrm{a}^{}_n>-b_{\mathrm{safe}}$.

\textbf{The incentive criterion} determines the advantage of HV after the lane change, quantified by the total acceleration gain, given by
\begin{equation}
\label{equ:mobilcondition}
\mathrm{a}'_{ego}-\mathrm{a}_{ego}+\sin \phi_{ego} \Big( (\mathrm{a}'_n-\mathrm{a}^{}_n) + (\mathrm{a}'_o-\mathrm{a}^{}_o) \Big) > \Delta a_{th}
\end{equation}
where $\mathrm{a}^{}_{o}$, $\mathrm{a}^{}_{n}$ and $\mathrm{a}^{}_{ego}$ represent the acceleration of the original follower in the current lane, the new follower in the target lane and the ego HV, correspondingly, and $\mathrm{a}'_{o}$, $\mathrm{a}'_{n}$, and $\mathrm{a}'_{ego}$ are the equivalent accelerations considering that the ego HV has changed the lane, $\sin \phi_{ego}$ is the politeness factor.
Finally, the lane change is performed if the safety and incentive criteria are mutually satisfied.

The IDM Model determines the longitudinal acceleration of a HV $\dot{v}_{\mathrm{k}}$ as follows,
\begin{equation}
\label{equ:idm1}
\dot{v}_{\mathrm{k}}=\mathrm{a}_\mathrm{max}\Big[ 1- \Big( \frac{v_k}{v_{\mathrm{k}}^0} \Big)^\delta - \Big( \frac{d^*(v_k, \Delta v_k)}{d_k} \Big)^2 \Big]
\end{equation}
in which $v_k$, $d_k$, $\delta$, $\Delta v_k$, $v_{\mathrm{0}}^k$ denote the speed, the actual gap, the acceleration exponent, the approach rate, and the desired speed of the $k^{th}$ HV, respectively. 

The desired minimum gap of the $k^{th}$ HV is given by,
\begin{equation}
\label{equ:idm2}
d^*(v_k, \Delta v_k) = d_k^0 +v_kT_\mathrm{k}^0 + \frac{v_k \Delta v_k}{ (2\sqrt{\mathrm{a}_{\mathrm{max}}.\mathrm{a}_{\mathrm{des}}})}
\end{equation}
where $T_k^0$, $d_k^0$, $\mathrm{a}_{\mathrm{max}}$, and $\mathrm{a}_{\mathrm{des}}$ are the safe time gap, the minimum distance, the comfortable maximum acceleration, and deceleration, correspondingly.

The typical parameters for the MOBIL model are
$\sin \phi_{ego}=0.5$, $\Delta a_{th} = 0.1 \frac{m}{s^2}$ 
and $b_{\mathrm{safe}} = 4 \frac{m}{s^2}$. Table~\ref{table:idm} shows typically used parameters of the IDM model~\cite{treiber2000congested}.

\begin{table}[h!]
\centering
\caption{\small{Common used parameters for the IDM model}}
\begin{tabular}{ c||ccccccc } 
 \textbf{Parameter} & $v^0$ & $T^0$ & $\mathrm{a}_{\mathrm{max}}$ & $\mathrm{a}_{\mathrm{des}}$& $\delta$ & $d^0$  \\
 \hline
  \hline
 \textbf{Value} & 30 m/s & 1.5 s & 1 m/s$^2$ & 1.5 m/s$^2$ & 4 & 2 m \\
 \hline
\end{tabular}\\[10pt]
\label{table:idm}
\end{table}

\textbf{Heterogeneous Driver Behaviors.}
Although those parameters are typically used for IDM and MOBIL models, they simulate just one behavior. In order to generate diverse behaviors $\mathcal{B}$, 
we frame the task of simulating diverse behaviors as the problem of obtaining the appropriate range of parameters ($\mathcal{P}$) that can generate those behaviors. To achieve that, we leverage a behavior classifier and iteratively simulate the parameters and classify the behaviors, mapping parameters to behaviors. 
To classify the behaviors we represent traffic using a traffic-graph at each time step $t$, $\mathcal{G}_t$, with a set of edges $\mathcal{E}(t)$ and a set of vertices $\mathcal{V}(t)$ as functions of time, i.e, the positions of vehicles ($ \widetilde{\mathcal{H}} \cup \widetilde{\mathcal{I}} $) represent the vertices. The adjacency matrix $A_t$ is given by $A(k,m) = d(v_k,v_m), k \neq m$ , in which $d(v_k,v_m)$ is the shortest travel distance between vertices $k$ to $m$. Then we use centrality functions~\cite{chandra2020cmetric} to classify the behavior (level of aggressiveness) resulting from $\mathcal{P}$, and then use those simulation parameters $\mathcal{P}$ to model behaviors within the simulator with varying levels of aggressiveness. The centrality functions are defined as,

\textbf{Closeness Centrality:} the discrete closeness centrality of the $k^\textrm{th}$ vehicle at time $t$ is defined as,
\begin{equation}
    \mathcal{C}^k_C[t] = \frac{{N-1}}{\sum_{v_m\in \mathcal{V}(t)\setminus \{v_k\}} d_t(v_k,v_m)},
    \label{eq: closeness}
\end{equation}
where $N = |\widetilde{\mathcal{H}} \cup \widetilde{\mathcal{I}}|$.
The more central the vehicle is located, the higher $\mathcal{C}^k_C[t]$ and the closer it is to all other vehicles.

\textbf{Degree Centrality:} the discrete degree centrality of the $k^\textrm{th}$ vehicle at time $t$ is defined as,
\begin{equation}
    \begin{aligned}
    \mathcal{C}^k_D[t] = \bigl | \{ v_m \in \mathcal{N}_k(t) \} \bigr | + \mathcal{C}^k_D[t-1] &\\
    \textrm{such that} \ (v_k,v_m) \not\in \mathcal{E(\tau)}, \tau = 0, \ldots, t-1&
    \end{aligned}
    \label{eq: degree}
\end{equation}

in which $\mathcal{N}_k(t) = \{ v_m \in \mathcal{V}(t), \ A_t(k,m) \neq 0, \nu_m \leq \nu_k\}$ represents the set of vehicles in the proximity of the $k^\textrm{th}$ vehicle, given that $\nu_m \leq \nu_k$; and $\nu_m, \nu_k$ denote the velocities of the $m^\textrm{th}$ and $k^\textrm{th}$ vehicles, $A_t(k,m)$ is the adjacency matrix. The more new vehicles seen by vehicle $k$ that meet this condition, the higher $\mathcal{C}^k_D[t]$.


With the centrality functions, we can measure the Style Likelihood Estimate (SLE) for different driver styles~\cite{chandra2020cmetric}. We consider two SLE measures. The SLE of overtaking and sudden lane changes ($SLE_l$) and the SLE of overspeeding ($SLE_o$).
The $SLE_l$ and $SLE_o$ can be computed by measuring the first derivative of the centrality functions as, 
\begin{equation}
   \textrm{SLE}_l(t) = \abs*{\frac{\partial \mathcal{C}_C(t)}{\partial t}} \quad
   \textrm{SLE}_o(t) = \abs*{\frac{\partial \mathcal{C}_D(t)}{\partial t}}
    \label{eq:sle}
\end{equation} 
The maximum likelihood 
$\textrm{SLE}_\textrm{max}$ is calculated as $\textrm{SLE}_{\textrm{max}} = \max_{t \in \Delta t}{\textrm{SLE}}(t)$.

Using those functions, we can approximately quantify and classify driver behaviors in our simulation. The intuition behind that is that an aggressive driver may frequently overspeed or perform sudden lane changes; while overspeeding the $\mathcal{C}_D(t)$ monotonically increases (higher $\textrm{SLE}_o(t)$) and during sudden lane changes the slope and the extrema of $\mathcal{C}_C(t)$ changes values. Thus higher values of $\textrm{SLE}_{\textrm{max}}$ are related to increased levels of aggressiveness. Conversely, conservative drivers are not inclined toward those aggressive maneuvers, and the degree of centrality will be relatively flat, 
thus $ \textrm{SLE}_o(t) \approx 0$ for conservative drivers.

We use these metrics as approximations of the driver's level of aggressiveness. In order to compute the suitable values for our simulation, we iteratively simulate the parameters from IDM and MOBIL models, and for each set of parameters, we quantify the resulting behavior in the simulation (using those metrics). A mapping of the parameters $\mathcal{P}$ to behaviors (quantified in the simulation for those parameters).
The estimated simulation parameters that simulate conservative, moderate and aggressive behavior in our scenarios are presented in Table~\ref{table:parameters}.

\begin{table}[htbp]
\caption{\small{Estimated simulation parameters for conservative, moderate, and aggressive behaviors.}}
\centering
\begin{tabular}{c|c|ccc} 

Model & Parameter & Aggressive  &  Moderate & Conservative \\
\hline
\hline
MOBIL & $\sin \phi_{e}$ & 0  & 0.3 & 1\\ 
& $\Delta a_{th}$ & 0 $m/s^2$ & 0.1 $m/s^2$ & 0.4 $m/s^2$ \\ 
& $b_{\mathrm{safe}}$ & 12.0 $m/s^2$ & 6.0 $m/s^2$ & 2.0 $m/s^2$\\ 

\hline
IDM & $T^0$ & 0.5s   & 1s & 3s \\ 
& $d^0$ & 1 $m$ & 2 $m$ & 6.0 $m$\\ 
& $\mathrm{acc}_{\mathrm{max}}$  & 7.0 $m/s^2$ & 3.0 $m/s^2$ & 1.0 $m/s^2$\\ 
& $\mathrm{acc}_{\mathrm{des}}$ & 12.0 $m/s^2$        & 7.0 $m/s^2$ & 2.0 $m/s^2$\\ 
\hline
\end{tabular}

\label{table:parameters}
\vspace{-10pt}
\end{table}

The desired velocity $v^0$ is set to $30m/s$ and the acceleration exponent $\delta = 4$.



\subsection{Implementation and Computational Details} 
\vspace{-0.1cm}
We customize the OpenAI Gym environment in~\cite{leurent2019approximate} to suit our particular driving situation and MARL problem. We design a merging ramp and exiting highway scenario for our simulation running in python and used Pytorch for the implementation of our safety prioritized MARL DDQN algorithm. Our implementation on average uses 3.1GB of memory for 4 agents and 18 HVs using a GPU NVIDIA Tesla V100. The training process is repeated several times to ensure convergence of the experiments to a similar policy. The network is trained for $N_{episodes} = 10,000$ taking on average 8 hours. While each round of $10,000$ training episodes in the Tesla V100 GPU takes around 8h, a full forward pass during deployment for 4 simulated agents takes 15ms (approximately 4ms per agent). We utilize 3,200 GPU hours for all our simulation experiments. Table~\ref{table: hyperparameters} lists our simulation and training hyper-parameters.

In a real AV platform, each agent will receive a local observation of the environment that will be used by our algorithm to compute the safe optimal action based on the trained Q-network. The decision-making will take place on each AV's onboard computer; therefore, to verify the feasibility of the real-time operation of our decentralized algorithm we tested a forward pass of the Q-network during deployment in multiple hardware platforms. The results for the different platforms are presented in Table~\ref{table:computation}, for instance, an online forward pass of the network in the deployment phase using commodity GPU hardware, i.e, an NVIDIA Jetson AGX platform will be around 32.9ms for each agent.

\begin{table}[h!]
\centering
\caption{\small{Computation time for each agent.}}
\begin{tabular}{ cc } 
  \hline
 \textbf{Computing platform} & Online forward pass time \\
 \hline
NVIDIA Tesla V100 GPU & 3.7 ms \\
OnLogic Karbo 700 x2 & 65.2 ms \\
NVIDIA Jetson AGX Xavier GPU & 32.9 ms \\
NVIDIA Jetson TX2 GPU & 112.5ms \\
 \hline
\end{tabular}\\[10pt]
\label{table:computation}
\end{table}
\vspace{10pt}
\begin{table}[t]
\caption{\small{List of hyper-parameters.}}
\begin{center}
\begin{tabular}{c c | c c}
Parameter &
Value &
Parameter &
Value\\ 
\hline
\hline
$N_{\mathrm{episode}}$ &
10,000 &
$\epsilon$ decay &
Linear  \\
$RM$ buffer size &
8,000 &
Initial exploration $\epsilon_0$ &
1.0 \\ 
Batch size & 
32 &

Final exploration & 
0.05 \\ 
Learning rate $\alpha_0$ &
0.0005 &
Optimizer &
ADAM \\ 
$Target_{update}$ &
300 &
Discount factor $\gamma$ &
0.95 \\ 
$|\mathcal{H}|$&
$18$ &
$|\mathcal{I}|$ &
$4$ \\ 
\hline
\end{tabular}
\end{center}
\label{table: hyperparameters}
\end{table}

\section{Experiments and Results} 
\label{sec:experiments}


\subsection{Manipulated Variables} 
We study how the $safe_{th}$, the \emph{level of aggressiveness}, the \emph{traffic scenarios} ($f_j$) and the \emph{HVs' behaviors} ($b_k$) impact the performance of AVs. We consider the case in which the mission vehicle (exiting/merging) in Figure~\ref{fig:mainfigure} is \emph{human-driven}, $M \in \mathcal{H}$, and define the following terms: 
\begin{itemize}
    \item \textbf{$AV_S$}. Social AV ($\phi_i = \phi^*$) that act \emph{altruistically} in the presence of diverse HVs behaviors $b \in \mathcal{B}$.
    \item \textbf{$AV_E$}. Egoistic AV ($\phi_i = 0$) that act \emph{egoistically} in the presence of diverse HVs behaviors $b \in \mathcal{B}$.
\end{itemize}

with $\phi^*$ to be the optimal SVO angle tuned to reach the optimal level of altruism as in~\cite{toghi2021social}.
\subsection{Performance Metrics}
The performance of our system is measured based on safety, efficiency, altruistic performance gain ($PG$), and adaptation error $\mathrm{A}_\mathrm{error}$. To measure safety, we compute the percentage of episodes that encountered a crash ($C(\%)$). For efficiency, the average traveled distance ($DT(m)$) of the vehicles and the number of missions accomplished by the mission vehicle is used. The altruistic performance gain is measured by computing the difference in the safety/efficiency performance of \textbf{$AV_E$} and \textbf{$AV_S$}, as
\begin{equation}
    PG_{safety}(\%) = \frac{(AV_E)_{C(\%)} - (AV_S)_{C(\%)}}{N_{Episodes}} 
\end{equation}

\begin{equation}
    PG_{efficiency}(\%) = \frac{(AV_S)_{DT(m)} - (AV_E)_{DT(m)}}{(AV_E)_{DT(m)}} 
\end{equation}
Finally, the adaptation error is a weighted sum function of the safety ($C(\%)$) and efficiency ($DT(m)$) performance of the \textbf{$AV_S$} when trained and tested in different scenarios/behaviors. Defined as
\begin{equation}
    A_{error}(\%) = w_{s}\times (C(\%)) + w_{e}\times 100(1-\frac{DT}{DT_{max}})
\end{equation}
such that an adaptation between different situations that result in $0\%$ crash and $DT = DT_{max}$ will have $\mathrm{A}_\mathrm{error}=0\%$. 

\vspace{-0.2cm}
\subsection{Hypotheses} 

In this section we examine the following hypotheses 
\begin{itemize}
    \item \textbf{H1.} \emph{In a mixed-autonomy scenario, the higher the level of aggressiveness, the bigger the impact of cooperation. We expect a higher performance gain ($PG$) when altruistic AVs face more aggressive environments}.
    \item \textbf{H2.} \emph{Altruistic AVs agents using the decentralized framework can adapt to different driver behaviors and traffic scenarios without compromising the overall traffic metrics. However, the higher the similarity of testing scenarios to the ones seen during training ($(f_{test},b_{test}) \approx (f_{train},b_{train})$), the lowest adaption error ($\mathrm{A}_\mathrm{error}$)}.
    \item \textbf{H3.} \emph{We anticipate an improvement in both safety and efficiency with the addition of the safety prioritizer. In the absence of a safety prioritizer ($safe_{th}=0$) we expect that AVs will cause more crashes}.

\end{itemize}


\subsection{Analysis and Results}
\label{sec:results}
Based on the hypotheses, we explore their correctness through the experiments in this section.

\subsubsection{Sensitivity Analyses} 
\begin{figure}[t]
  \centering
  \includegraphics[width=.7\textwidth]{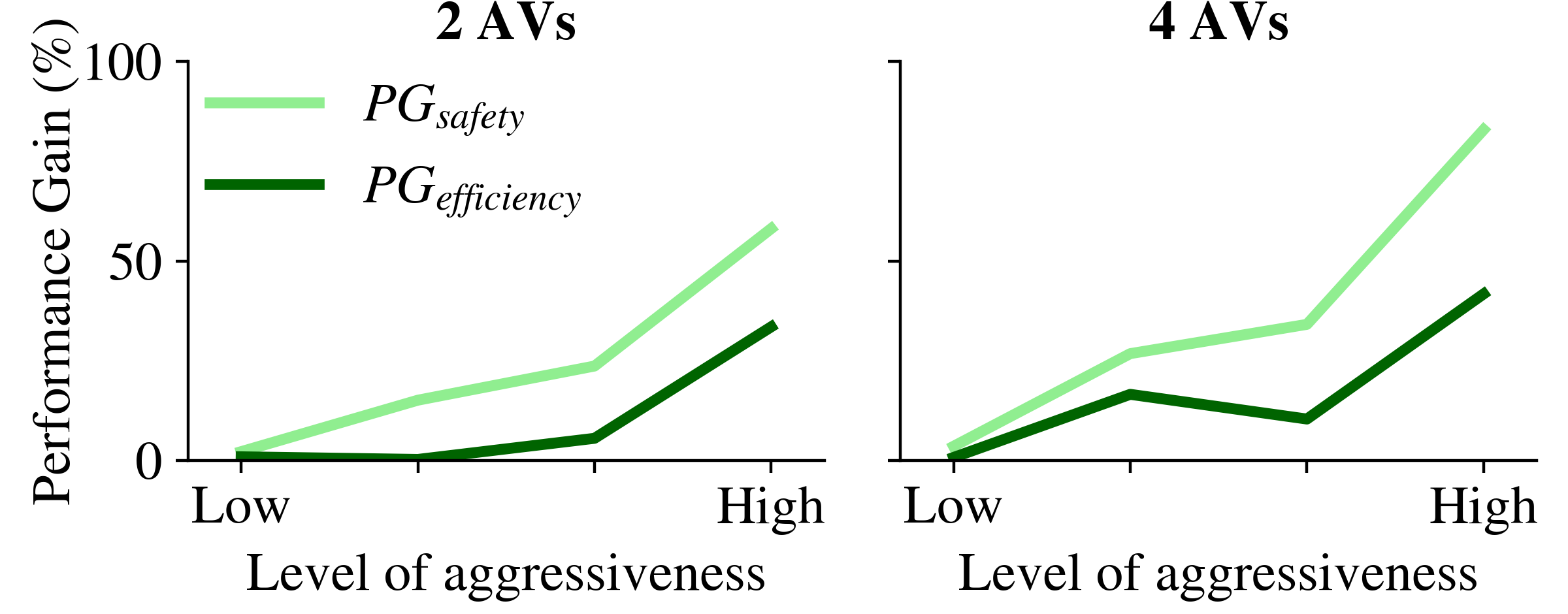}
  \caption{\small{Sensitivity analyses measured by altruistic performance gains (PGs) of AVs show that the more aggressive the HVs are, the more the impact/gain of cooperation.}}
  \label{fig:sensitivity1D}
\end{figure}

\begin{figure}[t]
  \centering
  \includegraphics[width=.7\textwidth]{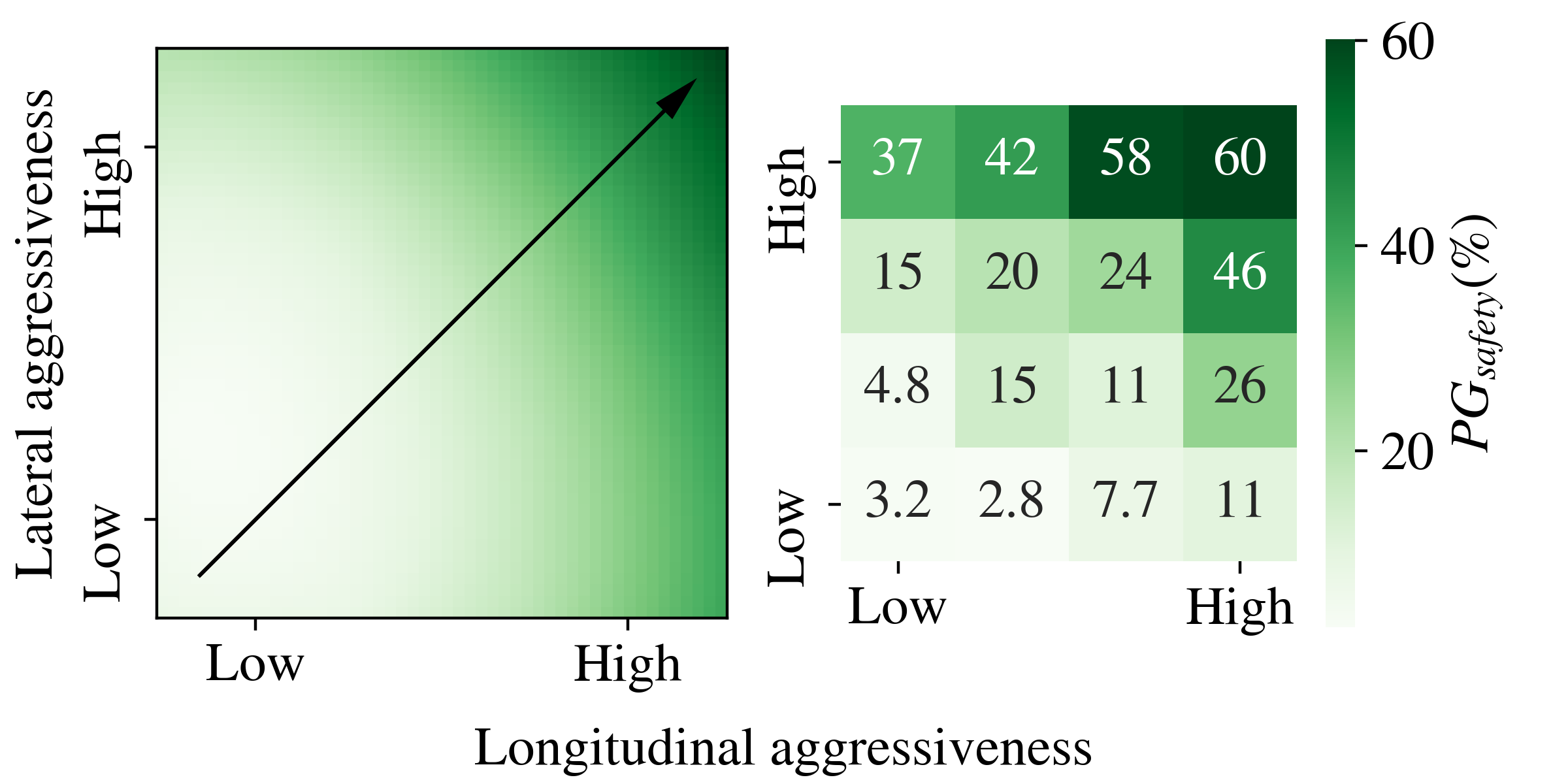}
  \caption{\small{Both lateral and longitudinal sensitivity analyses indicate an increase in altruistic performance gain (PG). 
  }}
  \label{fig:sensitivity2D}
\end{figure}

To study the hypothesis \textbf{H1} we investigate the effect of HV behaviors on the altruistic AV agents. We focus on scenarios with a HV mission vehicle, with safe AVs that act \emph{altruistically} ($AV_S$) or \emph{egoistic} ($AV_E$), in environments with increasing levels of HVs aggressiveness. 
Figure~\ref{fig:sensitivity1D} illustrates the altruistic performance gain for increasing levels of HVs' aggressiveness for 2 AVs (left) and 4 AVs (right). It demonstrates that the more aggressive the HVs are, the higher the impact of cooperation and thus confirms the \textbf{H1}.
This is also observed in Figure~\ref{fig:sensitivity2D} where the level of aggressiveness is decomposed into lateral and longitudinal aggressiveness. Lateral and longitudinal aggressiveness is varied by changing the MOBIL and IDM parameters (Table~\ref{table:parameters}) from aggressive to conservative. Figure~\ref{fig:sensitivity2D} shows that the altruistic gain increases in both directions, but is more pronounced in the longitudinal direction. That is probably due to the simulated scenarios having more longitudinal maneuvers.

\subsubsection{Domain Adaptation}
\label{sec:deepnetworks}
\begin{figure}[t]
  \centering
  \includegraphics[width=.7\textwidth]{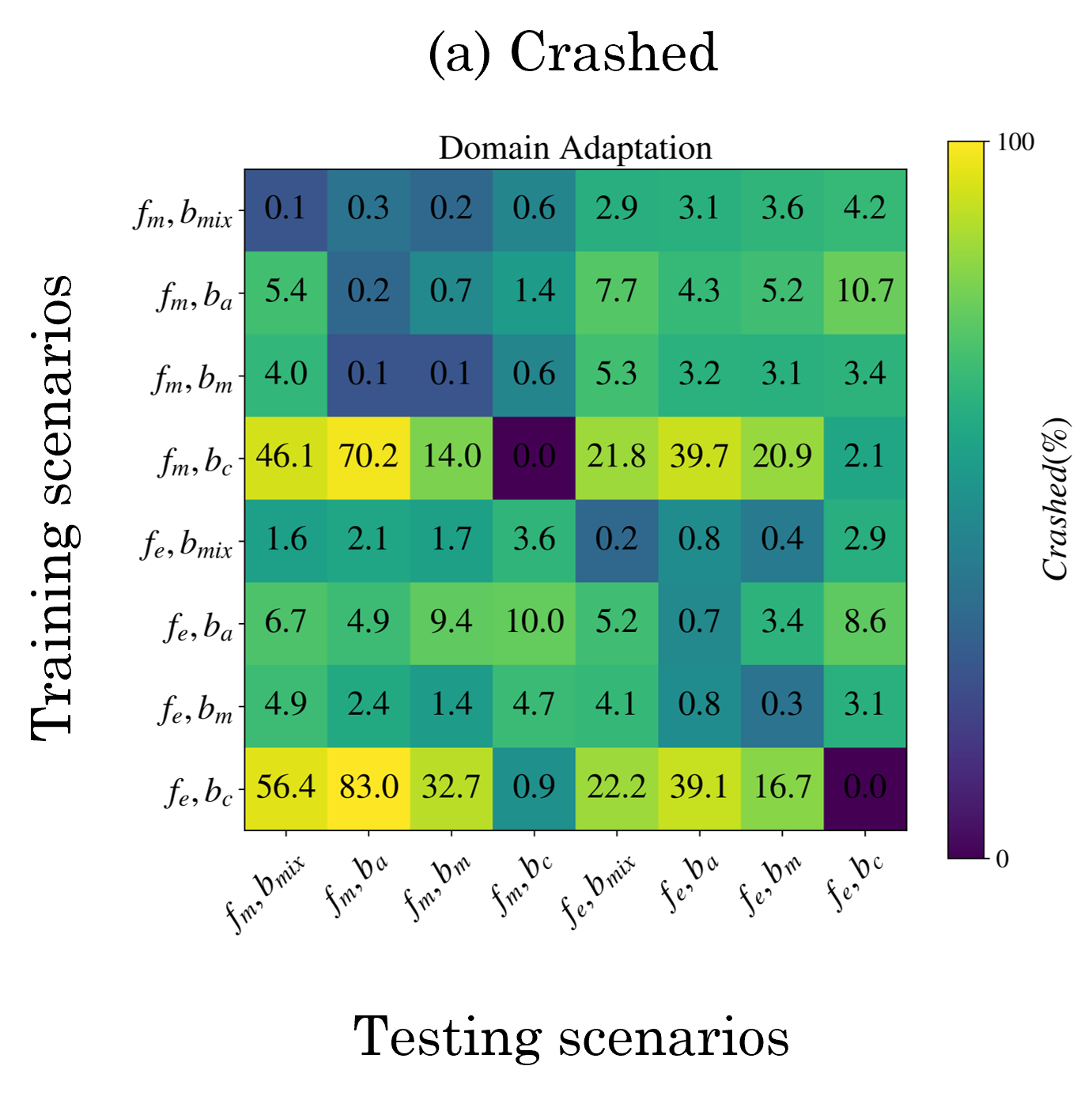}
  \caption{\small{\textbf{(a)} The domain adaptation matrix with crash percentage ($C(\%)$) between different traffic scenarios and behaviors. The lower $C(\%)$ the most suitable the adaptability in terms of safety (measured by $C(\%)$) between those domains.  $\mathrm{AV}_\mathrm{S}$ are trained (rows of the matrix) in different scenarios $f_i \in \mathcal{F}$ in the presence of HVs with different behaviors $b_k \in \mathcal{B}$ and tested (columns of the matrix) in other scenarios $f_j \in \mathcal{F}$ and behaviors $b_l \in \mathcal{B}$. Each pair ($f_i,b_k$) is a combination of scenario and behavior.}}
  \label{fig:adaptation_all}
\end{figure}

\begin{figure}[t]
  \centering
  \includegraphics[width=.7\textwidth]{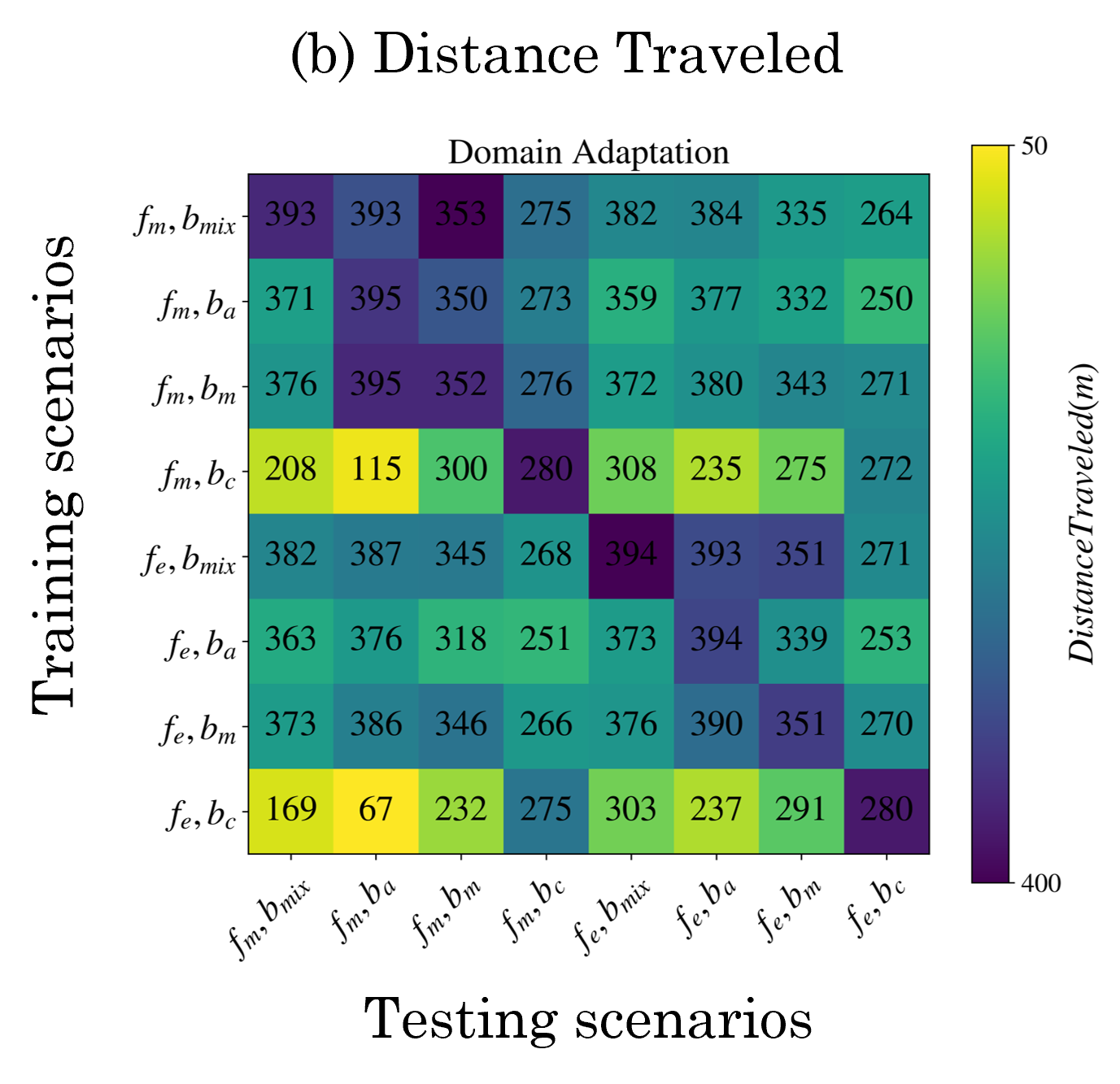}
  \caption{\small{\textbf{(b)} The domain adaptation matrix with distance traveled ($DT(m)$). Illustrating how the AVs adapt to other situations in terms of efficiency (measured by $DT(m)$).}}
  \label{fig:adaptation_all2}
\end{figure}

\begin{figure}[t]
  \centering
  \includegraphics[width=.7\textwidth]{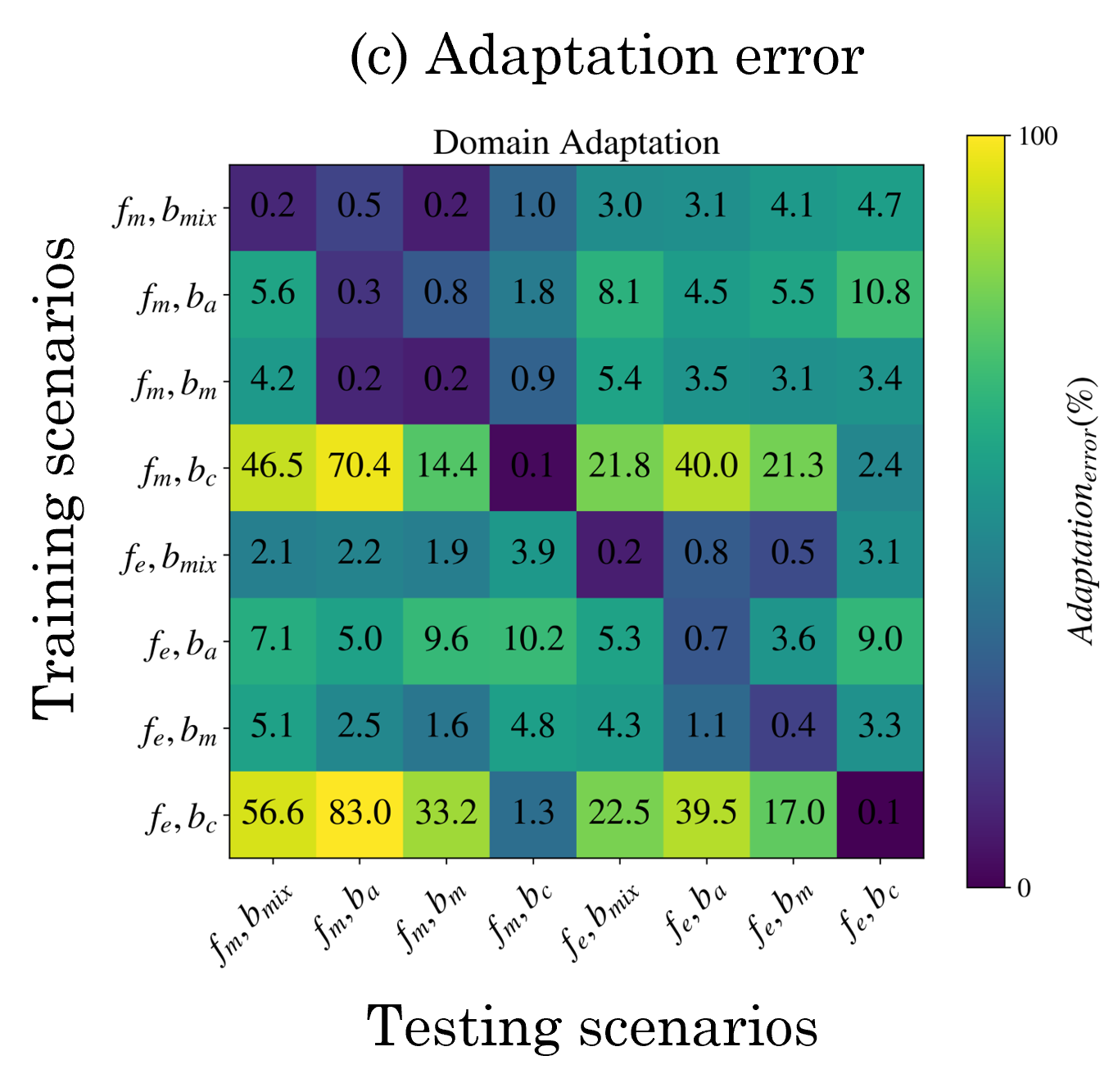}
  \caption{\small{\textbf{(c)} The domain adaptation matrix with adaptation error ($\mathrm{A}_\mathrm{error}$) between different traffic scenarios and behaviors. The lower $\mathrm{A}_\mathrm{error}$ the most suitable the adaptability between those domains.}}
  \label{fig:adaptation_all3}
\end{figure}

Following the sensitivity analysis, we investigate the domain adaptation of the AVs to validate the \textbf{H2}. Figure~\ref{fig:adaptation_all}, Figure~\ref{fig:adaptation_all2} and Figure~\ref{fig:adaptation_all3} show how the altruistic AVs learn to adapt to different scenarios and behaviors by different performance metrics, i.e, crashed \textbf{(a)}, distance traveled \textbf{(b)} and adaptation error \textbf{(c)}.
For the experiments, $AV_S$ are trained in different scenarios $f_i \in \mathcal{F}$ in the presence of HVs with different behaviors $b_k \in \mathcal{B}$ and tested in other scenarios $f_j \in \mathcal{F}$ and behaviors $b_l \in \mathcal{B}$. In our experiments, we consider two case study scenarios $f_e , f_m \in \mathcal{F} $ (exiting/merging) in environments with three different HVs behaviors $b_a, b_m, b_c \in \mathcal{B}$ (aggressive, moderate, conservative) see Table~\ref{table:parameters}; and a mixed behavior environment, in which HVs are created randomly and their behaviors are selected based on a uniform distribution over the behaviors in $\mathcal{B}$, given equal probability to the defined behaviors. In total, we have eight combinations of scenarios and behaviors, namely: ($f_m,b_{mix}$), ($f_m,b_a$), ($f_m,b_m$), ($f_m,b_c$), ($f_e,b_{mix}$), ($f_e,b_a$), ($f_e,b_m$), ($f_e,b_c$). 

The results are presented in Figure~\ref{fig:adaptation_all3} \textbf{(c)} as an adaptation matrix, showing the $\mathrm{A}_\mathrm{error}$ for different domains, the $\mathrm{A}_\mathrm{error}$ is in percentage ($\%$) and color-map in logarithmic scale to increase the perceived dynamic range for visualization. In our analyses, the weights used for $\mathrm{A}_\mathrm{error}(\%)$ are $w_{s} = \frac{2}{3}$ and $w_{e} = \frac{1}{3}$, which weighs the safety performance higher. $DT_{max}$ is computed based on the maximum distance for each situation. 
Additionally, Figure~\ref{fig:adaptation_all} \textbf{(a)} and Figure~\ref{fig:adaptation_all2} \textbf{b)}  illustrate how the AVs adapt in terms of safety (measured by $C(\%)$) and efficiency (measured by $DT(m)$), separately.

The matrix shows the best performances in its diagonal; where agents are trained and tested in the same environment (($f_i,b_k$); ($f_j,b_l$) with $i = j$ and $k = l$); due to the fact that agents experience similar situations during testing as they do during training. The vehicles trained in the merging environment can perform the exiting mission for different behaviors, and vice-versa. Interestingly, the performance of AVs trained in a conservative environment ($b_c$) is poor when tested in an aggressive environment ($b_a$). 
We believe that the reason is that in conservative environments, the HVs yield the mission vehicle, and the AVs learn to rely on HVs to guide the traffic. This learned policy is valid in a conservative environment where one can expect the HVs to always create a safe space for the mission vehicle. However, the same is not valid in more aggressive environments, in which AVs have to guide the traffic to avoid dangerous situations. As a result, the performance of vehicles trained in a conservative environment and tested in an aggressive one is the worse.

On the other hand, an adequate performance adaptation (lower $\mathrm{A}_\mathrm{error}$) is obtained when agents are trained in the presence of all moderate HVs ($b_m$) or a mixed behavior environment ($b_{mix}$), in which AVs face situations where the HVs yield, but also situations that require learning how to guide the traffic to optimize for the social utility. The results from the domain adaptation matrix indicate that a moderate or mixed environment is the most suitable for training robust AVs and show the adaptability of AVs to different situations, thereby confirming the \textbf{H2} hypothesis. 

It can be concluded that the adaptation between the environments is not reciprocal and environment and situations selection should be considered during training, based on the application needs and target situations. The Domain adaptation matrix identifies the settings in which altruistic AVs can best learn cooperative policies that are robust to different traffic scenarios and human behaviors.

\subsubsection{Transfer Learning}
\begin{figure}[t]
  \centering
  \includegraphics[width=.8\textwidth]{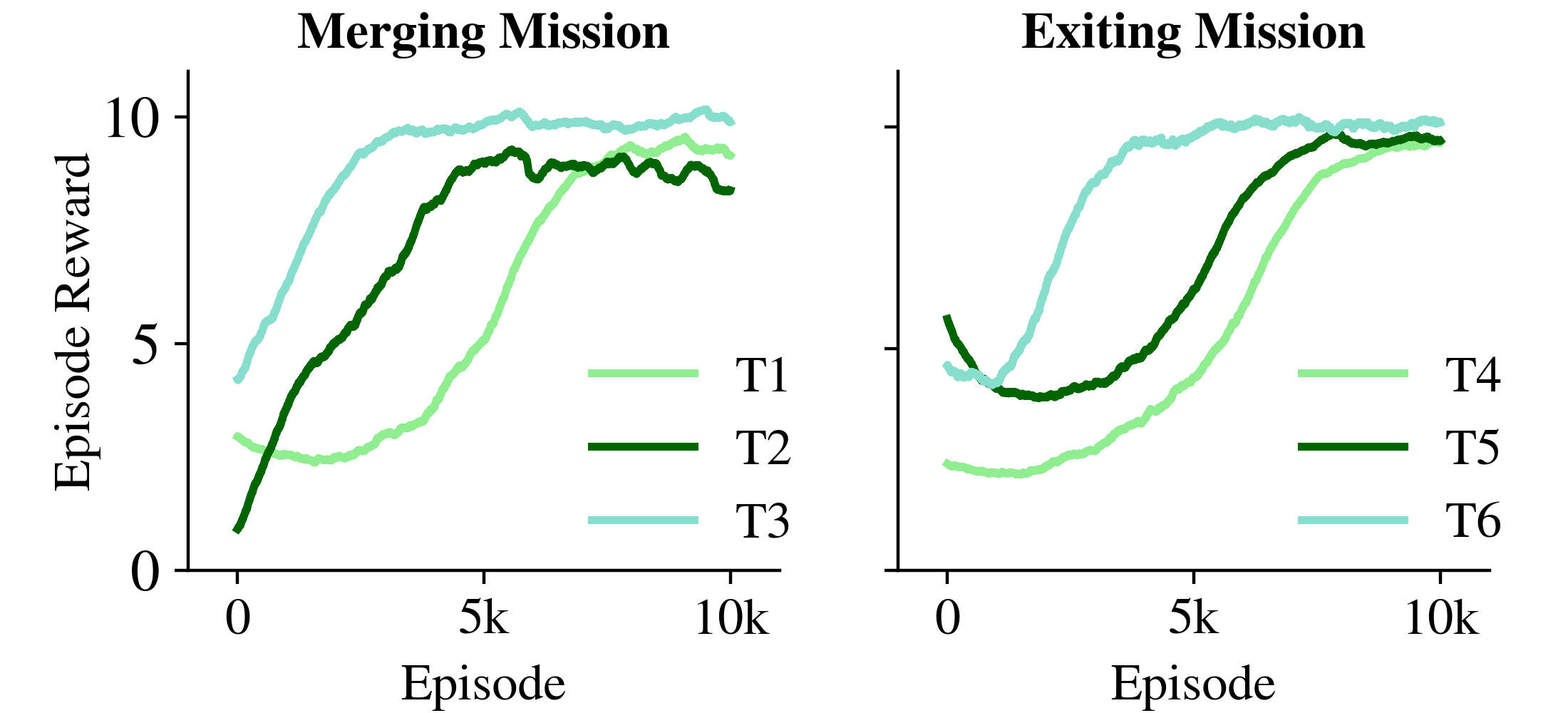}
    \caption{\small{The figure demonstrates how policies learned from merging can be transferred to the exiting environment to speed up the learning process while archiving similar performance to learning the task from scratch.}}
  \label{fig:transferlearning}
\end{figure}

Through domain adaptation and transfer learning, we promote generalization while learning harder tasks efficiently from trained models and accelerate the learning process. We study how the policies learned during merging can be transferred to the exiting environment. For that, we train AVs agents from scratch for the mission/task of merging $\mathrm{AV}_\mathrm{merging}$ (T1), train AVs agents to drive on a highway, and then use that model as the starting point to learn the merging task $\mathrm{AV}_\mathrm{drive-to-merging}$ (T2), train AVs agents for the exiting task and then use that model as the starting point to learn the merging task $\mathrm{AV}_\mathrm{exiting-to-merging}$ (T3); and apply the same procedure for the exiting task, learning to exit from scratch $\mathrm{AV}_\mathrm{exiting}$ (T4), after learned how to drive $\mathrm{AV}_\mathrm{drive-to-exiting}$ (T5) and after learned how to merge $\mathrm{AV}_\mathrm{merging-to-exiting}$ (T6). The results of the experiments are presented in Figure~\ref{fig:transferlearning} and show that our transfer learning approach speeds up the learning process while archiving similar performance as when learning the task from scratch. 

\subsubsection{Safety}
Finally, we compared state of the art architectures related to our approach~\cite{van2016deep,toghi2021social,toghi2021cooperative,toghi2021altruistic} in terms of safety and efficiency to validate \textbf{H3}. We trained the different architectures in the same situations and examined their performance under different levels of HVs behaviors. As noted in Table~\ref{table:comp} our safe altruistic agents consistently outperformed the other approaches, and the results are more notable when the level of aggressiveness is higher. We conclude that when using the safety prioritizer, immediate collisions are avoided reducing the overall number of crashes in the episodes.  Our agents can learn from scratch not only how to drive, but also to understand the behavior of HVs and coordinate with them.
\begin{table*}[t]
\caption{\small{A comparison of the performance of related architectures. Our safe altruistic AVs outperform the other solutions, and performance improvements become more noticeable as the level of aggressiveness increases.}}
\begin{center}
\begin{tabular}{c| c c c| c c c | c c c}
&
\multicolumn{3}{c}{Aggressive HVs} &
\multicolumn{3}{c}{Moderate HVs} &
\multicolumn{3}{c}{Conservative HVs}
\\
 \hline 
 \hline
Approaches &
C (\%)&
MF (\%) &
DT (m)&
C (\%)&
MF (\%) &
DT (m)&
C (\%)&
MF (\%) &
DT (m)\\
\hline
\hline
Conv2D+DQN~\cite{van2016deep}& 31.2 &
28.9 &
316 &
25.4 &
20.3 &
302 &
14.0&
7.9 &
274 \\

Toghi~\etal~\cite{toghi2021social}&
21.3 &
16.4 &
339 &
12.7&
10.1 &
333 &
1.6 &
0.6 &
269 \\

Conv3D+A2C~\cite{toghi2021altruistic} &
14.8 &
12.6 &
341 &
9.4 &
8.8 &
328 &
1.1 &
0.1 &
267 \\

Conv3D+DQN~\cite{toghi2021cooperative} &
3.1 &
2.8 &
359 &
2.6 &
2.4 &
341 &
0.3 &
\textbf{0} &
\textbf{284} \\
\textcolor{black}{\textbf{Ours}} & 
\textcolor{black}{\textbf{0.2}} & 
\textcolor{black}{\textbf{0.1}} & 
\textcolor{black}{\textbf{397}}& 
\textcolor{black}{\textbf{0.1}} & 
\textcolor{black}{\textbf{0.1}} & 
\textcolor{black}{\textbf{354}} & 
\textcolor{black}{\textbf{0}} & 
\textcolor{black}{\textbf{0}} & 
\textcolor{black}281\\
\hline
\end{tabular}
\end{center}
\raggedright\footnotesize{\hspace{0.4cm} \quad \quad \quad \quad \quad  C: \emph{Crashed}, MF: \emph{Mission Failed}, DT: \emph{Distance Traveled}}\\
\label{table:comp}
\end{table*}

\subsubsection{Importance of Social Coordination}
We demonstrate that social awareness and coordination are essential to improve safety and reliability on the roads. Particularly in our sensitivity analyses (Figure~\ref{fig:sensitivity1D}) we have shown that altruistic agents have a significant performance gain when compared to egoistic agents and the gain is more notable as the road becomes more aggressive. Additionally, to show that the performance gain vs. driver behaviors is not just because of a single altruistic agent but as a consequence of coordination among agents, we complement our results and conducted an experiment with the difference that only $\mathrm{AV1}$ is altruistic and the others are egoistic AVs, we label this scenario as single altruistic agent \textbf{SAA}. Table~\ref{table:nmultiagent} demonstrates the necessity of multi-agent coordination and the fact that a single altruistic AV, i.e., the Guide AV, is not able to achieve the mission of safe and seamless merging without help from the other AVs. Our results show that a non-cooperative SAA is not enough to guide the traffic and successful completion of the missions, as coordination is not guaranteed in a single-agent setting. All the AVs have to coordinate collectively to allow safe and efficient traffic, and this is unfeasible if the others do not collaborate. Table~\ref{table:nmultiagent} complements our results in Figure~\ref{fig:sensitivity1D} and support the hypothesis \textbf{H1}.

\begin{table}[htbp]
\caption{Importance of Social Coordination: AVs require to coordinate to enable a safe and seamless merging /exiting and none of them can achieve this goal if the others do not cooperate.}
\centering
\begin{tabular}{c|cccc} 

 &  Aggressive HVs $C (\%)$  &   Moderate HVs $C (\%)$   & Conservative HVs $C (\%)$  \\
\hline
\hline
\textbf{MAA} multi-agent altruistic  &  $0.2\%$  &  $0.1\%$     &  $0\%$  \\   
\textbf{SAA} single altruistic agent  &  $24.1\%$      &    $17.4\%$     &  $2.3\%$ \\

\hline
\hline
\end{tabular}

\label{table:nmultiagent}
\end{table}

\subsection{Qualitative Analyses} 

\begin{figure*}[t]
  \centering
  \includegraphics[width=.99\textwidth]{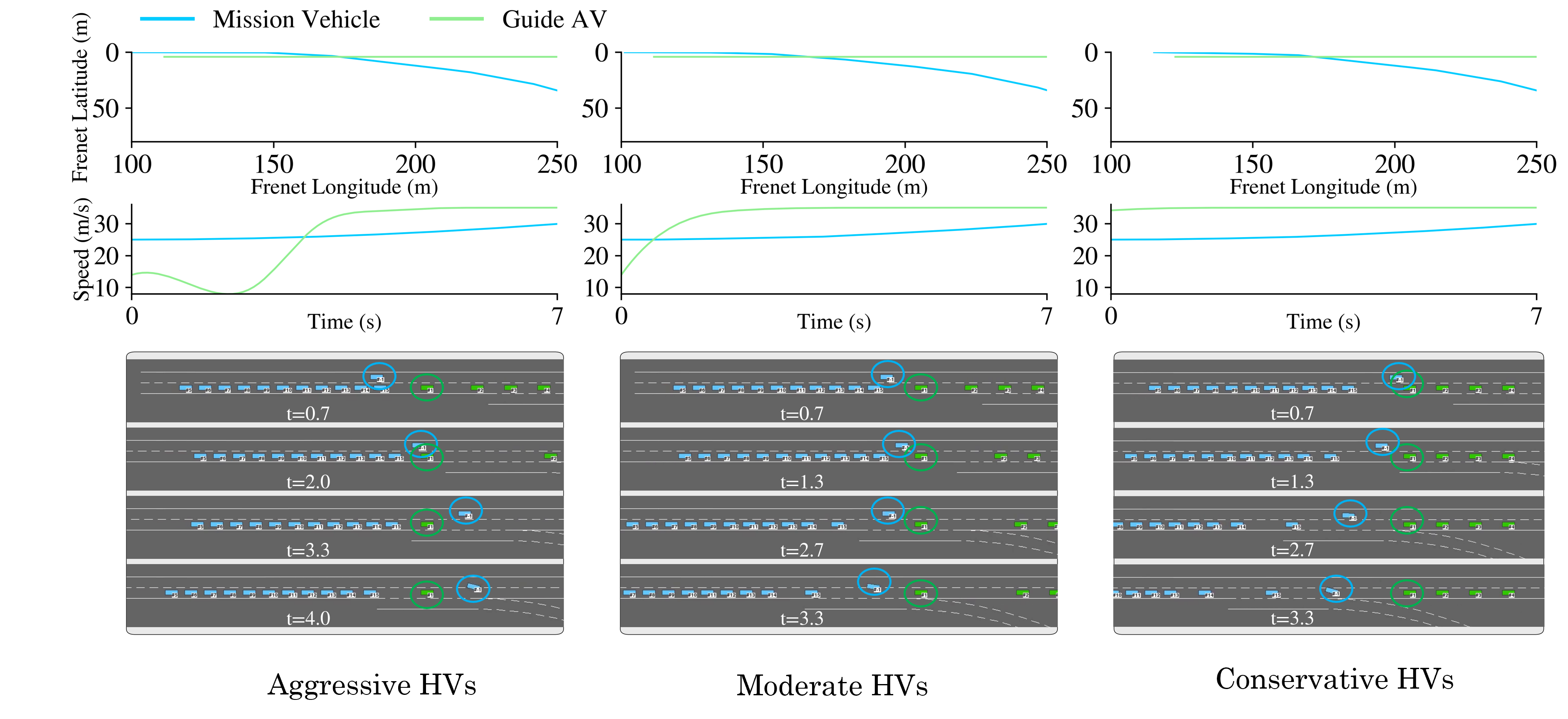}
  \caption{\small{Mission vehicle exiting the road under different HV behaviors (from left to right: aggressive, moderate and conservative HVs). AVs are shown in \textit{green} and HVs are shown in \textit{blue}.}}
  \label{fig:exitpolicy}
\end{figure*}

\begin{figure*}[t]
  \centering
  \includegraphics[width=.99\textwidth]{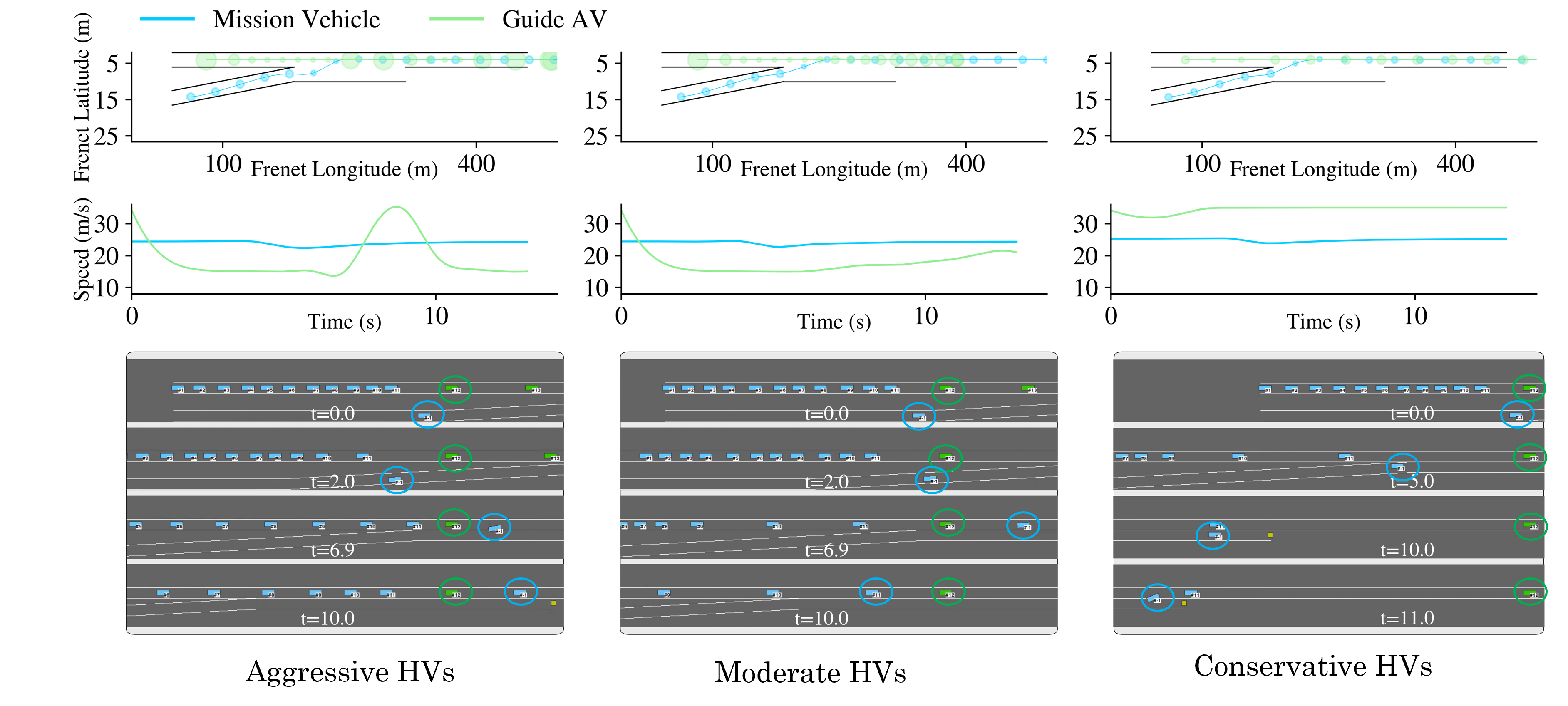}
  \caption{\small{Mission vehicle merging into the highway under different HV behaviors (from left to right: aggressive, moderate, and conservative HVs). AVs are shown in \textit{green} and HVs are shown in \textit{blue}. The diameter of the circles on the trajectory plot (first-row plot) shows the vehicles' speed.}}
  \label{fig:mergingbehaviors}
\end{figure*}

We show a qualitative analysis of our altruistic AVs in the exit and merging scenarios. Figure~\ref{fig:exitpolicy} provides further intuition about the policies learned by altruistic AVs (\textit{green}) in different situations, Figure~\ref{fig:exitpolicy} and  Figure~\ref{fig:mergingbehaviors} show a set of snapshots for different policies learned in an exit/merging environment in the presence of HVs (\textit{blue}) with different behaviors. In the presence of aggressive HVs, the guide AV has to slow down and guide the HVs in the platoon to allow a safe merging/exit of the mission vehicle; in this case, by slowing down the AV learn to compromise on their own utility for a more desirable social outcome. In the presence of moderate HVs behaviors, the guide AV slows down (slowing down the vehicles in the platooning) to open a safe space for the mission vehicle and then quickly accelerates, the space created by the quick AV intervention is safe enough to allow the mission vehicle to exit/merge the road; in this case, the AV compromise in their own utility but does not need to compromise as much as in the aggressive traffic scenario, it learns to take sequences of actions to not only enable the mission vehicle to merge (by quick decelerating), but also manages to make the minimum compromise on its individual utility.
Finally, in the conservative environment, the HVs are cautious enough to allow the mission vehicle to exit/merge safely, so the AVs learn to accelerate in those scenarios as the mission vehicle has enough space to merge, optimizing for their own utility (higher speed and longer distance travel), while also considering other vehicles utilities and safety; in this case the AV doe not need to compromise their own utility, it learns that HV will allow the exiting/merging so AVs does not need to guide the traffic. It is important to notice that the policies are learned by AVs from experience to optimize the social utility, AVs learn to adapt to different scenarios and behaviors. Is interesting to observe that our AVs develop some form of social awareness and learn the HVs' behaviors from experience, acting accordingly to optimize traffic efficiency while prioritizing safety.

\section{Concluding Remarks}
\label{sec:concluding}

AVs need to learn to co-exist with HVs vehicles as deploying egoistic AVs that solely account for their individual interests on the road leads to sub-optimal and non-desirable social outcomes. Social awareness and coordination are essential to improve safety and reliability on the roads.  We demonstrate how altruistic AVs learn the decision-making process from experience, considering the interests of all vehicles while prioritizing safety and optimizing a general decentralized social utility function. We expose the settings for our MARL problem in which transfer learning and domain adaptation are more feasible, and conducted a sensitivity analysis under different HVs' behaviors.
Our experiments reveal that altruistic AVs learn to leverage social coordination to improve safety and reliability. Our social-aware AVs are robust to heterogeneous driver behaviors and can form alliances and affect the behavior of HVs to create socially-desirable outcomes that benefit the group of the vehicles.

\smallskip
\textbf{Future Work:} 
Although we explored various elements of social navigation in a variety of settings and the presence of diverse HV behaviors, the HV models used are not from real human driver data, and the traffic scenarios are limited to merging and exiting. However, we believe that by leveraging and learning from actual human data and traffic circumstances, our approach might be beneficial in practical traffic conditions. For this strategy to be used in real-world circumstances, more attention to safety is necessary.
We intend to investigate more sophisticated architectures and state representations in future work, as well as develop a more realistic simulation environment that incorporates data from real-world traffic and can handle more complex interactions between HVs and AVs, as well as diverse traffic agents like bicycles and pedestrians. Despite the drawbacks, we are excited to see safe and reliable social-aware AVs on the road that learns from experience. Beyond driving, we expect these principles to be applied to general multi-agent human-robot interactions in which agents influence humans and collaborate safely for a socially beneficial result.

\bibliographystyle{IEEEtran}
\bibliography{IEEEbibs} 
\end{document}